\documentclass[conference]{IEEEtran}
\IEEEoverridecommandlockouts
\usepackage{cite}
\usepackage{amsmath,amssymb,amsfonts}
\usepackage{algorithmic}
\usepackage{graphicx}
\usepackage{textcomp}
\usepackage{xcolor}
\def\BibTeX{{\rm B\kern-.05em{\sc i\kern-.025em b}\kern-.08em
    T\kern-.1667em\lower.7ex\hbox{E}\kern-.125emX}}
    
\usepackage{booktabs}
\usepackage{numprint}    
\usepackage{multirow}    
\usepackage{subcaption}
\usepackage{hyperref}

\begin{document}

\title{R+R: Understanding Hyperparameter \\ Effects in DP-SGD}

\author{\IEEEauthorblockN{Felix Morsbach}
\IEEEauthorblockA{\textit{KASTEL Security Research Labs} \\
\textit{Karlsruhe Institute of Technology}\\
Karlsruhe, Germany \\
felix.morsbach@kit.edu}
\and
\IEEEauthorblockN{Jan Reubold}
\IEEEauthorblockA{\textit{KASTEL Security Research Labs} \\
\textit{Karlsruhe Institute of Technology}\\
Karlsruhe, Germany \\
jan.reubold@kit.edu}
\and
\IEEEauthorblockN{Thorsten Strufe}
\IEEEauthorblockA{\textit{KASTEL Security Research Labs} \\
\textit{Karlsruhe Institute of Technology}\\
Karlsruhe, Germany \\
thorsten.strufe@kit.edu}
}

\maketitle

\begin{abstract}
Research on the effects of essential hyperparameters of DP-SGD lacks consensus, verification, and replication. 
Contradictory and anecdotal statements on their influence make matters worse.
While DP-SGD is the standard optimization algorithm for privacy-preserving machine learning, its adoption is still commonly challenged by low performance compared to non-private learning approaches. 
As proper hyperparameter settings can improve the privacy-utility trade-off, understanding the influence of the hyperparameters promises to simplify their optimization towards better performance, and likely foster acceptance of private learning.

To shed more light on these influences, we conduct a replication study: We synthesize extant research on hyperparameter influences of DP-SGD into conjectures, conduct a dedicated factorial study to independently identify hyperparameter effects, and assess which conjectures can be replicated across multiple datasets, model architectures, and differential privacy budgets.
While we cannot (consistently) replicate conjectures about the main and interaction effects of the batch size and the number of epochs, we were able to replicate the conjectured relationship between the clipping threshold and learning rate.
Furthermore, we were able to quantify the significant importance of their combination compared to the other hyperparameters. 
\end{abstract}


\section{Introduction}

Replication studies serve the research by verifying experimental results, by refining scientific theories, and ultimately by helping establish highest levels of reliability in the scientific record of knowledge. 
Therefore, where any line of inquiry encounters a problem of scattered and unverified insights, replicatory works can help streamline understanding and so advance the research. 
One such problem is found in the recent research on hyperparameter effects in differentially private stochastic gradient descent (DP-SGD). 
Multiple recent works have contributed propositions and valuable insights into these effects. 
However, the insights are scattered across many works \cite{abadi_deep_2016, mcmahan_general_2018, mcmahan_learning_2018, vanderveen_three_2018, bagdasaryan_differential_2019, papernot_making_2019, papernot_hyperparameter_2022, tramer_differentially_2021, li_large_2022, dormann_not_2021, de_unlocking_2022, kurakin_toward_2022} and what is more, the insights are not consistently and reliably based on independent empirical verification. 
Here especially, replication can create consensus and refine the current body of knowledge.

DP-SGD \cite{abadi_deep_2016} has become the de facto standard for training machine learning models with differential privacy guarantees \cite{yousefpour_opacus_2021, andrew_tensorflow_2018}. 
Yet, models trained with DP-SGD perform worse than their counterparts trained with stochastic gradient descent (SGD) \cite{jayaraman_evaluating_2019}.
By properly adjusting the learning pipeline, recent works not only managed to significantly decrease the accuracy gap to SGD.
They also demonstrated that hyperparameter effects differ between SGD and DP-SGD. 
For example, multiple works demonstrated that the optimal batch size is significantly larger in DP-SGD than in SGD \cite{dormann_not_2021, li_large_2022}, and further observations of other hyperparameter effects have been made.
However, to date, no systematic or replicatory studies have been conducted on the hyperparameter effects in DP-SGD.

In this work, we conduct a replication study on how DP-SGD's essential hyperparameters (i.e. batch size, number of epochs, learning rate, and clipping threshold) affect model accuracy on non-convex machine learning tasks.
We do not intend for this work to break ground on state-of-the-art performance; therefore, we also do not consider  specific alterations to the learning pipeline \cite{remerscheid_smoothnets_2022, de_unlocking_2022} or pre-training on public data \cite{de_unlocking_2022, li_large_2022}.
Instead, we focus on refining the understanding of the essential hyperparameters' effect on DP-SGD.
To this end we review extant literature which discusses the effect of hyperparameters of DP-SGD or otherwise reports on them. 
Based on these insights we synthesize six testable conjectures. 
To assess the replicability of the conjectures, we conduct a dedicated experiment to independently identify the hyperparameter effects of DP-SGD in a systematic and structure way. 
As most research on the effect of hyperparameters of DP-SGD considered image classification and language modeling, we also chose these two domains for our study.
In our factorial experiments we evaluate $3822$ hyperparameter tuples across six datasets, six model architectures, and three differential privacy budgets. 
Based on the results, we assess the hyperparameters' importance, their main, and interaction effects.  
Subsequently, we discuss the effects identified in our experiment in relation to the conjectured effects from related work.
Besides enabling us to assess the replicability\footnote{We follow the terminology of the ACM on replicability, see \url{https://www.acm.org/publications/policies/artifact-review-and-badging-current}.} of conjectures from related work, this large-scale experiment also provides the most comprehensive investigation on the hyperparameter effects of DP-SGD to date.

In summary, we make the following contributions:
\begin{itemize}
\item In a first step toward replication, we review the literature and collect scattered insights into hyperparameter effects in DP-SGD. Further, we synthesize these insights into six testable observations about hyperparameter effects of DP-SGD.
\item Our large-scale factorial study across multiple datasets, model architectures, and differential privacy budgets independently identifies and quantifies the importance of main and interaction effects of the four essential hyperparameters of DP-SGD.
\item We evaluate which of the synthesized conjectures about the hyperparameter effects in DP-SGD can be supported by independent experimental results.
\end{itemize}

The remainder of this paper is organized as follows:
In Section \ref{sec:background} we introduce DP-SGD, how the differential privacy budget is calculated, and how the budget interacts with the algorithm's hyperparameters.
In Section \ref{sec:related_work} we review the related work on the hyperparameters effect and synthesize their insights into conjectures.
In Section \ref{sec:study} we motivate and introduce our study design including what machine learning tasks we evaluated, how we chose and evaluated the hyperparameter space, and especially, how we identified main and interaction effects based on the results from our experiment.
In Section \ref{sec:results} we report the identified main and interaction effects of the hyperparameters and discuss the replicability of the conjectures from related work. 
In Section \ref{sec:conclusion} we conclude this study.

\section{Background}
\label{sec:background}
A randomized mechanism $\mathcal{M}\colon \mathcal{D} \rightarrow \mathcal{R}$ with domain $\mathcal{D}$ and range $\mathcal{R}$ satisfies $(\varepsilon,\delta)$-differential privacy \cite{dwork_algorithmic_2014} if for any two adjacent inputs $d, d' \in \mathcal{D}$ and for any subset of (measurable) outputs $\mathcal{S} \subseteq \mathcal{R}$ it holds that 
\begin{equation*}
\Pr\left[\mathcal{M}(d) \in \mathcal{S}\right] \leq e^\varepsilon\,\Pr[\mathcal{M}(d') \in \mathcal{S}] + \delta.
\end{equation*}

To make the stochastic gradient descent (SGD) algorithm ($\varepsilon, \delta)$-differentially private, the most common approach requires two major alterations to the SGD algorithm \cite{abadi_deep_2016}.
First, for each step of SGD, the gradients of each training record inside a minibatch are individually clipped such that their $L_2$-norm does not exceed the clipping threshold $C$: 
Let $x$ denote a minibatch of training records of size $B$ that is sampled from all training records, $x_i$ denote the $i^{\textrm{th}}$ training record inside the minibatch, and $g(x_i)$ denote the calculated gradient for $x_i$.
In DP-SGD, the gradient $g(x_i)$ is scaled to $\overline{g}(x_i) \leftarrow g_t(x_i) / \text{max}(1, \frac{\|g_t(x_i)\|_2}{C})$. 
Note that this clipping is done before averaging the gradients over all examples in the minibatch.
Second, after all the individual and clipped gradients $\overline{g}(x_i)$ have been averaged together, Gaussian noise with variance $\sigma^2C^2$ (where $\sigma$ is called noise multiplier) is added to the averaged gradient: $\hat{g}(x) \leftarrow \frac{1}{B} (\sum_i \overline{g}(x_i)+Z)$ with $Z\sim\mathcal{N}(0,\sigma^2C^2I)$.
The clipped, noised, and averaged gradient $\hat{g}(x_i)$ is then used for a standard gradient descent step.

The differential privacy budget ($\varepsilon$) of training a model with DP-SGD is given by the batch size ($B$), the noise multiplier ($\sigma$), the number of epochs ($E$) and the number of training examples ($N$) used for training\footnote{For a more detailed derivation of how exactly these parameters define the differential privacy budget, please see Abadi et al. \cite{abadi_deep_2016} or McMahan et al. \cite{mcmahan_general_2018}.}.
Given values for these four hyperparameters and $\delta$, the privacy budget ($\varepsilon$) can be computed without the need to actually train the model. 
Increasing the number of epochs ($E$) or the batch size ($B$) increases the consumed privacy budget. 
Increasing the noise multiplier ($\sigma$) or the number of training examples ($N$) decreases the consumed privacy budget.
The clipping threshold ($C$), further hyperparameters, and model architecture choices do not influence the calculation of the differential privacy budget. 
Equation~\ref{eps_correlation} demonstrates this relationship (i.e. whether $\varepsilon$ increases or decreases with respect to the parameters).

\begin{equation}
\varepsilon\,(B, E, \sigma, N) \approx \dfrac{E, B}{\sigma, N}
\label{eps_correlation}
\end{equation}

However, as one usually targets a specific differential privacy budget, the noise multiplier $\sigma$ is not set a priori as Equation \ref{eps_correlation} would suggest, but calculated based on the other values:
Given a batch size $B$ and the dataset size $N$, the noise multiplier $\sigma$ is calculated such that the targeted privacy budget $\varepsilon$ is reached after training for $E$ epochs.
Consequently, when targeting a specific differential privacy budget, changing one of the three variables will require adjusting at least one of the remaining ones. 
For example, given a hyperparameter configuration, increasing the number of epochs in that configuration will either require decreasing the batch size or increasing the noise multiplier in order to achieve the same differential privacy budget.

While increasing the amount of training data ($N$) is not only beneficial to machine learning in general, it also decreases the consumed privacy budget of training a machine learning model with DP-SGD.
However, increasing the amount of training data is difficult, especially when working with sensitive data. 
We therefore do not further discuss this parameter in more detail, as one usually has no option to increase the amount of training data.

Please note that compared to SGD, the role and influence of hyperparameters in DP-SGD is slightly different and at times becomes more complex. 
For example, as also discussed by Dörmann et al. \cite{dormann_not_2021}, in classic SGD the batch size defines the amount of inherent sampling noise in each gradient step. 
In DP-SGD, however, it also affects the noise multiplier $\sigma$ via the privacy amplification by sub-sampling theorem (in combination with the dataset size) \cite{abadi_deep_2016}. 
Thus, by changing the batch size in DP-SGD, one does not only affect the inherent sampling noise of the minibatch as in SGD, but also the added noise of the Gaussian mechanism required to satisfy differential privacy. 
Further, the role and influence of the number of epochs and the learning rate also changes. 
Thus a direct comparison to the hyperparameter effects in SGD is non-trivial. 

\section{Related work}
\label{sec:related_work}
Many works discussed the effects of the hyperparameter on model accuracy in varying degrees of detail, but the insights are distributed across many works. 
Recently, Ponomareva et al. \cite{ponomareva_how_2023} (partially) summarized results on the influence of hyperparameters on DP-SGD. 
However, their summary does not contain any empirical evaluations of their validity. In contrast, we not only summarize previous findings from literature, but also test whether we can replicate these results with an independent and dedicated study. 
Further, we are also the first to additionally quantify the strength of the hyperparameters main and interaction effects by applying a functional analysis of variance.

In the following, we discuss the insights from related work on the effects of the essential hyperparameters (i.e. batch size, number of epochs, learning rate, and clipping threshold) on model accuracy and privacy budget. 
We focus our attention on observations that can be made from the results of the related work, the authors' observations, and the authors' conjectures made on the influence of the hyperparameter of DP-SGD. 
Where possible, we directly extract the conjectures made by related work or otherwise synthesize the observations into conjectures ourselves. In total, we form six groups labeled Conjecture 1 (C1) through Conjecture 6 (C6).

\subsection{Batch Size}
\label{sec:batch_size}
Increasing the batch size ($B$) will increase the consumed privacy budget (see Equation \ref{eps_correlation}), as larger batch sizes benefit less from privacy amplification by sub-sampling \cite{abadi_deep_2016}. 
Thus, intuitively, one would opt for a very small batch size to reduce the consumed privacy budget.
However, it has been shown empirically that larger batch sizes are preferable to smaller batch sizes in terms of maximizing model accuracy, even though increasing the batch size has to be balanced with an increase of the added noise or a decrease in the number of epochs. 

\subsubsection*{C1: The batch size is the most important hyperparameter}
\label{sec:hp-importance}
Besides considering the value of hyperparameters, some hyperparameters are also deemed more important than others.
Especially the batch size is deemed the most important hyperparameter to tune in DP-SGD \cite{li_large_2022} and its tuning should be preferred to adjusting other hyperparameters such as the clipping threshold or the noise multiplier \cite{mcmahan_general_2018}.
However, extant research on DP-SGD does neither operationalize this concept nor quantifies it empirically. 

\subsubsection*{C2: Increasing the batch size increases accuracy}
Finding an optimal batch size is discussed in many works and the general consensus is that larger batch sizes are better than smaller batch sizes for training with DP-SGD; for private training from scratch \cite{abadi_deep_2016, vanderveen_three_2018, bagdasaryan_differential_2019, mcmahan_general_2018, tramer_differentially_2021, de_unlocking_2022, dormann_not_2021} and also for private fine-tuning \cite{li_large_2022, kurakin_toward_2022, anil_largescale_2021}. 
However, while many works seem to agree on the concept of \textit{large} batch sizes being beneficial, the range of values tagged as \textit{large} can vary a lot.
For example, on CIFAR-10 and with varying models and hyperparameter settings the reported optimal batch sizes range from $1024$ to $16384$ \cite{tramer_differentially_2021,abadi_deep_2016,dormann_not_2021,de_unlocking_2022}.
While varying greatly, these proposed batch sizes are all significantly larger for DP-SGD than for SGD \cite{keskar_largebatch_2017}.

Dörmann et al. \cite{dormann_not_2021} offer a possible explanation for why larger batch sizes are beneficial. 
Stochastic gradient descent contains inherent sampling noise due to the minibatch sampling technique. 
This has been shown to be beneficial for generalization performance \cite{keskar_largebatch_2017}.
Since DP-SGD adds additional Gaussian noise onto the gradients of each training step,  D\"{o}rmann et al. argue that the inherent sampling noise of small batch sizes is no longer necessary to improve generalization performance.
Thus, large batch sizes paired with correspondingly large noise multipliers can lead to the same effect.
While D\"{o}rmann et al. do not explicitly state that this phenomenon is the cause of why larger batch sizes are beneficial for DP-SGD\footnote{They only state that increasing the noise multiplier and the batch size should increase generalization performance.}, it is a reasonable and interesting theory derivable from their work.

\subsubsection*{C3: The batch size has interaction effects}
There also have been many additional observations or recommendations for setting the other hyperparameters in conjunction with the batch size:
For example, Papernot et al. \cite{papernot_making_2019} and Dörmann et al. \cite{dormann_not_2021} suggest combining large batch sizes with few number of epochs to reduce the computational costs while retaining comparative performance.
Further, large batch sizes should be combined with a large learning rate \cite{vanderveen_three_2018} and its optimal value depends on the learning rate \cite{tramer_differentially_2021} and number of epochs \cite{li_large_2022, de_unlocking_2022}. 

\subsection{Number of Epochs}
In the non-private SGD setting, the number of epochs usually is not tuned as a hyperparameter.
Rather, a model is typically trained until convergence or until a predefined early-stopping criterion is met.
With DP-SGD, however, each epoch of training consumes privacy budget. 
This makes also intuitive sense:
Training for more epochs means that each training record is used for more updates to the model. 
But as in a differential privacy setting every data access consumes privacy budget, more epochs also consume more budget.
Thus, increasing the number of training epochs ($E$) will increase the privacy budget (see Equation \ref{eps_correlation}).

Towards choosing the number of epochs, recall that the noise multiplier is calculated such that the training reaches the desired privacy budget exactly after training for the specified number of epochs (given a batch size and number of epochs).
Thus, the optimal number of epochs would be such that the model has converged exactly after training for these number of epochs and that the privacy budget would have been exactly consumed.
Training for more epochs is not possible without exceeding the specified privacy budget and converging too early wastes privacy budget that would have been better spent elsewhere.

Therefore, the choice of the number of epochs is important and its misconfiguration can lead to unnecessary poor performance.

\subsubsection*{C4: The number of epochs has a significant effect on model accuracy}
Compared to the batch size, there are less insights on the optimal setting of the number of epochs, but multiple works attribute a measurable effect on model accuracy to the number of epochs.
De et al. \cite{de_unlocking_2022} claim that there exist an optimal value for the number of epochs and that as the batch size increases, the optimal number of epochs also seems to increase \cite{ponomareva_how_2023}.
Their results also indicate that the higher the targeted privacy budget, the more training steps are optimal (which is either achieved by decreasing the batch size or increasing the number of epochs).
Kurakin et al. \cite{kurakin_toward_2022} suggest that increasing the number of epochs and offsetting the additional privacy cost with a higher noise multiplier increases model accuracy.
Papernot et al. \cite{papernot_making_2019} suggest favoring a small number of epochs and spending the privacy budget on large batch sizes instead.
However, we deem there not to be an overarching theme that allows us to formulate a more precise conjecture.

\subsection{Clipping Threshold}
The clipping threshold ($C$) is the last hyperparameter that relates to the DP modifications of SGD.
While not directly impacting the privacy budget, it scales the amount of noise introduced and impacts the signal-to-noise ratio of the gradient updates. 
A higher clipping threshold will clip less gradients, therefore preserving more information.
However, the amount of noise per parameter scales linearly with the clipping threshold.
Thus, a higher clipping threshold increases the amount of added noise.
In contrast, a low clipping threshold clips the gradients more aggressively, reducing the amount of information in each gradient update, but also decreases the added noise. 

\subsubsection*{C5: The clipping threshold's effect is affected by the learning rate}
Multiple works described the influence of the clipping threshold in DP-SGD. 
Kurakin et al. \cite{kurakin_toward_2022} observed that the best private accuracy is obtained when the clipping threshold is chosen to be below a threshold value $C_T$.
Below that threshold, large clipping thresholds have been reported to pair well with low learning rates and vice versa \cite{kurakin_toward_2022, de_unlocking_2022, li_large_2022}.

\subsection{Learning Rate}
Even though the learning rate ($lr$) has no direct impact on the differential privacy budget of DP-SGD, it is an integral component of SGD.
Its setting is important for achieving reasonable performance \cite{goodfellow_deep_2016}. 
It has been shown empirically that the optimal value for the learning rate differs between SGD and DP-SGD \cite{papernot_making_2019, papernot_tempered_2021, tramer_differentially_2021, de_unlocking_2022}. 
However, it is inconclusive whether the optimal learning rate for DP-SGD is lower \cite{tramer_differentially_2021} or higher \cite{papernot_making_2019, papernot_tempered_2021, li_large_2022} than for SGD.

\subsubsection*{C6: The learning rate's effect is affected by the clipping threshold}
Kurakin et al. \cite{kurakin_toward_2022} observe a relationship between the clipping threshold and learning rate.
They observe that the highest model accuracy is achieved with inversely proportional learning rate and clipping threshold values (which would mean that the product $lr \cdot C$  is constant).
Kurakin et al. \cite{kurakin_toward_2022} conclude that there is an optimal update size such that values for the learning rate and the clipping threshold do not matter significantly, as long as their product is constant.

\section{Study Design}
\label{sec:study}
The primary goal of this study is to assess how well conjectures about the effects of DP-SGD's hyperparameters can be replicated across different datasets, model architectures, and differential privacy budgets.
However, the insights from related work, on which these conjectures are based, come from very diverse experimental settings, which often were not designed to derive hyperparameter effects in the first place.
For example, some insights are based on the results of a hyperparameter optimization run, which can cause a bias in the hyperparameter space, thus making these results often unfit for identifying hyperparameter effects \cite{moosbauer_explaining_2021}.
Other insights are derived from one-factor-at-a-time (OFAT) experiments, which cannot detect interaction effects and are prone to identify incorrect main effects \cite{montgomery_design_2017}.
Furthermore, some insights are based on experiments that do not consider multiple datasets, model architecture, or differential privacy budgets, making it unclear whether and how well the effects generalize. 
In consequence we abandon all experimental setups of original works. 
Instead, we identify the hyperparameter effects of DP-SGD independent of those works so that we may compare our new results to the conjectures derived from the experiments of the related works. 
For that reason, we conduct a factorial study to derive in a systematic and structure way both the main and interaction effects of hyperparameters in DP-SGD across multiple datasets, model architectures, and privacy budgets.

\subsection{Machine Learning Tasks, Datasets and Models}
To identify the effects of the hyperparameters on model accuracy, we conducted a fractional factorial study \cite{montgomery_design_2017}.
The four key hyperparameters of DP-SGD (batch size, number of epochs, clipping threshold, and learning rate) are independently randomly sampled, a model is trained with these hyperparameter values using the Opacus framework \cite{yousefpour_opacus_2021} as implementation of DP-SGD, and the models' test accuracy is recorded as a response variable. 
It is important to note that to discover the effects properly, the experiments have to follow a factorial design compared to the often-popular OFAT design. 

For the remainder of this paper, we define a scenario as tuple of dataset, model architecture and targeted privacy budget.
To cover as many scenarios used in the related works as possible while still keeping the computational complexity (and thus resource consumption) of factorial experiments in check, we chose to conduct three different sets of experiments. 
Each experiment set is a collection of multiple datasets, model architectures and privacy budgets.
We classify the experiment sets as a simple image classification task, an intermediate image classification task, and a simple text classification task (see Table \ref{tab:sets}). 
Inside each experiment set, the scenarios are full factorial and the hyperparameters are uniformly randomly sampled. 
The hyperparameter space is the same for every experiment set.

\begin{table}[]
    \centering
    \caption{Overview of the three experiment sets (IS = Image Simple, II = Image Intermediate, TS = Text Simple), datasets, architectures, privacy budget ($\varepsilon$), and number of hyperparameter tuples sampled ($\#S)$.}
    \begin{tabular}{ccccc}
    \toprule
    Set & Datasets & Architecture & $\varepsilon$ & $\#S$ \\
    \midrule
    IS & \{C10, SVHN\} & \{DP-CNN, R18, R34\} & \{3, 5, 7.5\} & 125 \\
    II & \{IN, C100\} & \{DP-CNN, R18, D121\} & \{7.5\} & 150 \\
    TS & \{NEWS, IMDB\} & \{RNN, LSTM\} & \{7.5\} & 168 \\
    \bottomrule
    \end{tabular}
    \label{tab:sets}
\end{table}

For the simple image classification set, we chose SVHN \cite{netzer_reading_2011} and CIFAR-10 \cite{krizhevsky_learning_2009} as comparatively small and easy datasets.
We chose two ResNets (i.e. R18 and R34) as popular and canonical vision architecture. 
To make the ResNets compliant with DP-SGD, we replaced the batch normalization layer with group normalization. 
Further, we included the DP-CNN architecture of Papernot et al. \cite{papernot_tempered_2021} because it outperforms conventional image processing architectures such as plain ResNets without additional architectural modifications or data augmentation at the chosen privacy budgets \cite{de_unlocking_2022}. 
We trained the models at three commonly used privacy budgets $\varepsilon \in \{3, 5, 7. 5\}$ with $\delta = 10^{-5}$.
For each of the $18$ scenarios, we evaluated $125$ uniformly randomly drawn hyperparameter tuples. 

For the intermediate image classification set, we chose more difficult datasets with ImageNette and CIFAR-100. 
ImageNette is a subset of the popular ImageNet dataset, containing \numprint{13393} images of size $160\times160$\,px across $10$ classes, compared to for example CIFAR-10's RGB images of size $32\times32$\,px \cite{howard_imagenette_2019}.
Additionally, we also included CIFAR-100 as it contains RGB images of $100$ different classes compared to CIFAR-10's $10$ classes.
Both are considered more difficult than CIFAR-10 or SVHN. 
As model architectures we used DP-CNN, a ResNet-18, and additionally a DenseNet-118 \cite{huang_densely_2017}.
As we did neither expect nor observe any meaningful difference between the privacy budgets in the simple image classification set, we only trained models with a targeted privacy budget of $\varepsilon = 7.5$ for the intermediate image and simple text set to reduce the amount of necessary computation.
For each of the six scenarios, we evaluated $150$ uniformly randomly drawn hyperparameter tuples.

To also investigate the hyperparameter effects in machine learning tasks from domains different to image classification, we included a text based domain, the second most popular domain for previous hyperparameter studies of DP-SGD. 
We choose the IMDB \cite{maas_learning_2011} and NEWS \cite{zhang_characterlevel_2015} datasets for text classification.
We used two simple model architectures which embed the words in $\mathbb{R}^{100}$ using GloVe \cite{pennington_glove_2014}, followed by an RNN or LSTM layer of size $192$, followed by a fully connected layer of size $96$ into the output layer, and use tanh as activation function.  
For each of the four scenarios, we evaluated $168$ uniformly randomly drawn hyperparameter tuples.

\begin{table}
\caption{Overview of the hyperparameter ranges tested in our experiment; a range is described by its lower and upper bound; $N$ denotes the number of data samples in a dataset.}
\label{tab:overview-hp-ranges}
\centering
\begin{tabular}{lccc}
\toprule
\multirow{2}{*}{Hyperparameter} & \multirow{2}{*}{Symbol} & \multicolumn{2}{c}{Range}  \\
& & From & To \\
\midrule
Batch Size & $B$ & $16$ & $N$ \\
Epochs & $E$ & $25$ & $500$ \\
Learning Rate & $lr$ & $10^{-5}$ & $10$ \\
Clipping Threshold & $C$ & $10^{-5}$ & $10$ \\
\midrule
Privacy budget & $\varepsilon$ & \multicolumn{2}{c}{$\{3$, $5$, $7.5\}$} \\
Relaxation param. & $\delta$ & \multicolumn{2}{c}{$\left\{10^{-5}\right\}$} \\
\bottomrule
\end{tabular}
\end{table}

\subsection{Hyperparameter Space}
One key aspect during the design of this study was to minimize the risk of missing well-performing regions in the hyperparameter space, ultimately skewing the conclusions on the hyperparameter effects. 
To mitigate this risk, we used a simple heuristic to evaluate the search space of each hyperparameter. 
After running the experiments, we partition the results to only include the hyperparameter tuples that perform well. 
We then examine the marginal density distribution of each hyperparameter in this subset. 
If the mode of this distribution is near the boundary of the chosen range and the search space could be extended in that direction, the chosen search space is insufficient and needs to be adjusted. 
Based on this heuristic, we adjusted the range of the number of epochs twice.
This led to the following hyperparameter space.

As the range of possible batch sizes is naturally limited by $1$ and the number of available training examples, recommendations regarding the optimal batch size vary widely in related work, and even full batch training being recommended, we did not limit the search range for the batch size. 
Thus, it ranges between $16$ (to still be able to benefit from GPU acceleration) and the size of the dataset (i.e. full batch training).
Based on observations from prior work, the range of the learning rate and the clipping threshold is set to be between $10^{-5}$ and $10$ \cite{kurakin_toward_2022} and the range for the number of epochs to be between $25$ and $500$ \cite{de_unlocking_2022}.
Table \ref{tab:overview-hp-ranges} summarizes the hyperparameter search space of the experiment.

\subsection{Identifying Main and Interaction Effects}
The training of a deep learning model can be modeled as a function evaluation: The learning algorithm (e.g. SGD) is a function that is parameterized by multiple variables (e.g. the hyperparameters, the model architecture, and the dataset) and returns a trained model (or, in our viewpoint, a model accuracy). Thus, the model accuracy becomes a function of the hyperparameters, model and dataset.

One way to characterize such a functional relationship between variables is by describing the main and interaction effects of the independent variables (i.e. the hyperparameters) on the dependent variable (i.e. the model accuracy). 
A main effect describes how the change in one independent variable will change the value of the dependent value, in our case how a single hyperparameter directly impacts model accuracy independently of and across all values of the other hyperparameters. 
An interaction effect describes how the effect of one independent variable on the dependent variable is affected by the value of another independent variable. 
Last, the strength of the effects -- the importance of (sets of) hyperparameters -- can be described via the amount of variance of the dependent variable they explain.

To identify the effects based on the results from a factorial experiment, one can fit a regression model to approximate the relationship between independent and dependent variables. 
Subsequently, the effects of the independent variables can be extracted from the fitted regression model. 
Explicating main and interaction effects from simple parametric regression models such as linear regression is straightforward, as the effects are the learned coefficients of each regressor. 
Unfortunately, linear regression is too simple of a model to properly approximate the relationship in question, as the relationship between the variables is not linear. 
Instead and based on an automated model selection and hyperparameter optimization run, we chose an extremely randomized trees regression model \cite{geurts_extremely_2006}. 
However, while tree-based models can approximate the relationship significantly better than linear models, identifying main and interaction effects from tree-based regression models is more complicated. 

One option is to use interpretable machine learning (IML) methods that try to explain the effects of features on the prediction of machine learning models. 
First, we analyzed the regression model with a functional analysis of variance (fANOVA) \cite{sobol_sensitivity_1993, hutter_efficient_2014} to assess the importance of each hyperparameter. 
For each input variable of a regression model (here any combination of DP-SGD's four hyperparameters), fANOVA calculates the amount of variance in the model's outcome (here predicted accuracy) the input explains. 
This not only works for single variables, but also for sets of input variables and thereby allows us to quantify the importance of the interaction effects.
For interaction effects, fANOVA produces two different values, the individual and the total amount of variance the set of input variables explain. 
The total variances describe the amount of variance that the combination of multiple input variable explain. It is the sum of the variance each input variable exclusively explains plus the amount variance the interaction individually explains. 
The individual variance of a set of variables describes the additional amount of variance that can only be explained by observing input variables in conjunction instead of in isolation. 
For example, let $V_1$ and $V_2$ be two variables, let $\operatorname{TI}$ denote the total importance of a set of variables and $\operatorname{II}$ denote the individual importance of a set of variables. Then $\operatorname{TI}(V_1, V_2) = \operatorname{TI}(V_1) + \operatorname{TI}(V_2) + \operatorname{II}(V_1, V_2)$.

Second, to assess the main and interaction effects we use independent conditional expectation (ICE) \cite{goldstein_peeking_2015} and accumulated local effect (ALE) \cite{apley_visualizing_2020} plots.
Both, ICE and ALE plots aim to describe how the output of a regression model is affected by an input along its range. 
Much simplified, they report the expected outcome of the regression model for each value of an input variable. 
Additionally, we used centered ICE (c-ICE) and derivative ICE (d-ICE) plots to make the analysis of interaction effects easier. 
In c-ICE plots the level effects of the individual conditional expectation curves are removed, thus effectively making them all start at the same vertical point. This is helpful to detect and identify interaction effects. 
The d-ICE plots display the first derivative of the individual conditional expectation curves, revealing whether interaction effects are equally distributed across their range.
As the fANOVA also describes the individual importance of sets of hyperparameters, thus the amount of variance their interaction explains, the fANOVA results informed our analysis towards which interaction effects are worth investigating. 

\paragraph*{Regression models for our analysis}
We fitted three extremely randomized tree regression models \cite{geurts_extremely_2006} -- one for each experiment set -- that approximate the relationship between batch size, number of epochs, learning rate, clipping threshold, dataset, model architecture, privacy budget, and test accuracy.
To make the results comparable across scenarios, we min-max scaled the test accuracy to be in $[0, 1]$ and further included dataset, model architecture, dataset size, and the sampling rate (i.e. the batch size divided by the dataset size) into the set of predictor variables.

For the majority, there was no difference in the effects between dataset, model architecture, and differential privacy budget, thus, we grouped all scenarios of one set together into one regression model to ease the subsequent analysis.
This aligns with theory, that predicts hyperparameter effects to be similar on similar scenarios (i.e. inside the experiment sets) \cite{vanschoren_meta-learning_2019, feurer_hyperparameter_2019}.
Thus, in the following analysis, we generally omit the absence of differences and only report if significant differences between dataset, model architectures, or differential privacy budget were noticeable.
The regression models had a mean absolute error on a holdout set between $0.036$ and $0.057$ on the min-max scaled test accuracy. 
We deem this to be accurate enough to derive general main and interaction effects.
For the subsequent analysis the regression models were refitted with all available data including the holdout set.

\section{Results}
\label{sec:results}
In the following we systematically analyse our results to determine the hyperparameters' importance, their main effects as well as their interaction effects. 
Subsequently and where applicable, we compare our findings with the conjectures derived from related work and discuss whether we were able to replicate results that support the conjectures.

\subsection{Hyperparameter Importance}
As first analysis step, we calculate the importance of each (set of) hyperparameter using a functional analysis of variance, see Table \ref{tab:HP_importance}. 
The results show that the learning rate ($lr$) and clipping threshold ($C$) are the most important hyperparameters. Individually they each already account for between $23\%$ and $28\%$ of the variance, their interaction for between $10\%$ and $13\%$, and combined for between $59\%$ and $60\%$.

In contrast, the batch size ($B$) and the number of epochs ($E$) seem rather unimportant overall, as they individually only account for $3\%$ or less of the total variance for the image classification task. There also does not seem to be a strong interaction between these two.

The other non-negligible interaction effects seem to be between the number of epochs and the learning rate ($E$+$lr$) or clipping threshold ($E$+$C$) respectively with $3.5\%$ and $4\%$ of accounted variance respectively for the simple image classification set.

\begin{table}
    \centering
    \caption{The amount of individual and total variance each hyperparameter subset explains across the three experiments. Values with non-negligible individual importance are highlighted in bold.}
    \begin{tabular}{ccccccc}
    \toprule
    \multirow{2}{*}{Hyperparameter} & \multicolumn{2}{c}{Image Simple} & \multicolumn{2}{c}{Image Inter.} & \multicolumn{2}{c}{Text Simple} \\
     & Indiv. & Total & Indiv. & Total & Indiv. & Total \\
    \midrule
    $B$ & 0.004 & 0.004 & 0.017 & 0.017 & 0.011 & 0.011\\
    $E$ & 0.026 & 0.026 & 0.007 & 0.007 & 0.001 & 0.001\\
    $lr$ & \textbf{0.232} & 0.232 & \textbf{0.232} & 0.232 & \textbf{0.281} & 0.281\\
    $C$ & \textbf{0.24} & 0.24 & \textbf{0.241} & 0.241 & \textbf{0.226} & 0.226\\
    $B$+$E$ & 0.002 & 0.032 & 0.003 & 0.026 & 0.001 & 0.012\\
    $B$+$lr$ & 0.009 & 0.245 & 0.01 & 0.259 & 0.018 & 0.309\\
    $E$+$lr$ & \textbf{0.035} & 0.292 & 0.016 & 0.255 & 0.005 & 0.287\\
    $B$+$C$ & 0.006 & 0.25 & 0.016 & 0.273 & 0.016 & 0.252\\
    $E$+$C$ & \textbf{0.04} & 0.306 & 0.011 & 0.259 & 0.004 & 0.231\\
    $lr$+$C$ & \textbf{0.116} & 0.588 & \textbf{0.128} & 0.6 & \textbf{0.096} & 0.602\\
    $B$+$E$+$lr$ & 0.003 & 0.31 & 0.003 & 0.287 & 0.001 & 0.318\\
    $B$+$E$+$C$ & 0.005 & 0.323 & 0.003 & 0.298 & 0.001 & 0.259\\
    $B$+$lr$+$C$ & 0.01 & 0.617 & 0.019 & 0.663 & 0.014 & 0.661\\
    $E$+$lr$+$C$ & 0.032 & 0.72 & 0.016 & 0.651 & 0.007 & 0.62\\
    $B$+$E$+$lr$+$C$ & 0.003 & 0.762 & 0.003 & 0.726 & 0.001 & 0.682\\
    \bottomrule
    \end{tabular}
    \label{tab:HP_importance}
\end{table}

\subsection{Main Effects}
In order to identify main effects we used individual conditional expectation (ICE) and accumulated local effect (ALE) plots. Please note that for clarity, the ICE plots only contain a random fraction of all ICE curves. 

\subsubsection{Batch size}
Analyzing the effect of the batch size on model accuracy has to be prepended with a word of caution:
Comparing or aggregating the batch size's effect across datasets is challenging as its range of possible values can be different for different datasets. 
For example, ImageNette has only 9469 training images while the NEWS dataset has 120k training examples. Thus, comparing their effect based on absolute values is difficult. Instead, analyzing the batch size as sampling rate (batch size divided by the number of training examples) seems to allow for an easier comparison as they are uniformly distributed in the input space across datasets.
However, one has to be keep in mind that the same sampling rates on two different datasets can translate to different actual batch sizes. 
As both approaches have benefits and drawbacks, we decided to incorporate both viewpoints into our analysis.

\begin{figure*}[!h]
     \centering
     \subfloat{\includegraphics[width=0.5\textwidth]{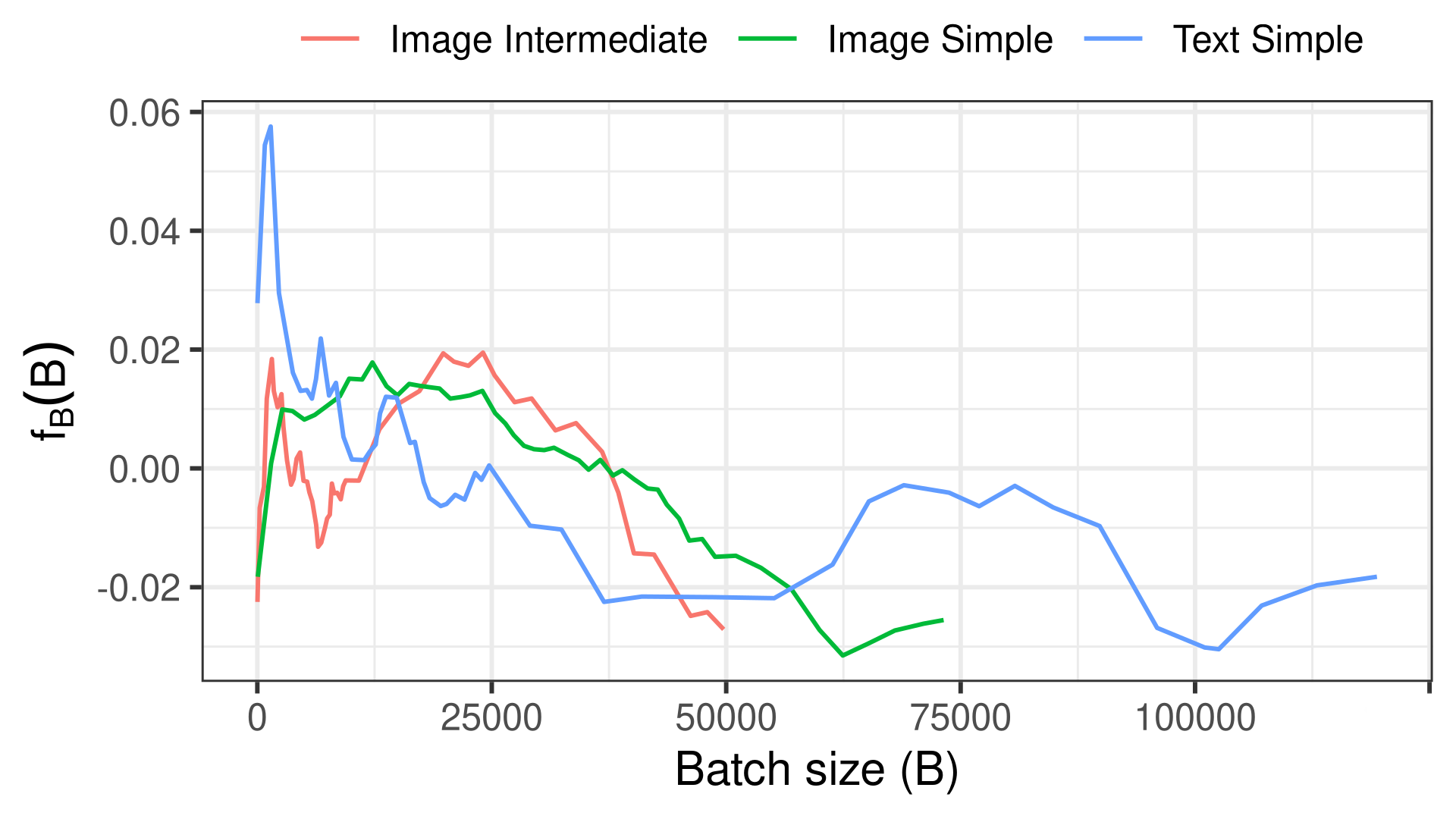}}
     \subfloat{\includegraphics[width=0.5\textwidth]{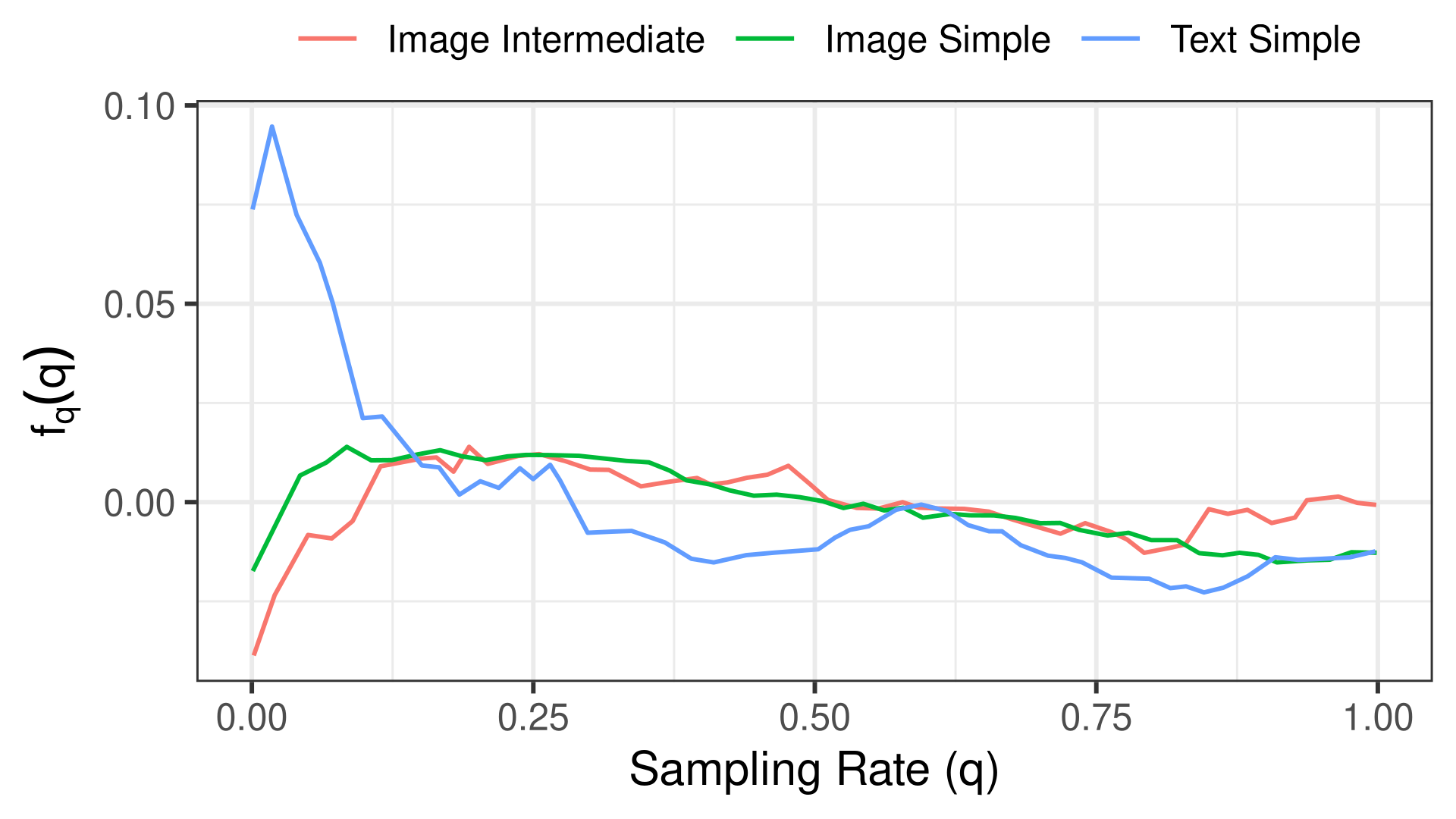}}
     \caption{ALE-plots for batch size (left) and sampling rate (right) across the three experiment sets}
    \label{fig:ale_batch_size_sampling_rate}
\end{figure*}

As already indicated by the fANOVA, the batch size’s effect on the expected min-max normalized model accuracy is not very strong: For the image classification tasks, the ALE plots reveal the batch size only affects the expected min-max normalized accuracy by at most~4\%, see Figure~\ref{fig:ale_batch_size_sampling_rate}.
Furthermore, the effect itself, even though its amplitude is very small, varies across dataset and model architectures. 
For example, in the simple image classification set, the batch size’s and sampling rate’s effect on the expected min-max normalized model architecture on CIFAR-10 plummets for batch sizes smaller than 10k while on SVHN the effect peaks in this area. 
Similarly, in the advanced image classification set, the batch size's effect with the DenseNet or ResNet peaks for batch sizes smaller than 5k, whereas the effect for the DP-CNN peaks at round 20k and is the lowest below 10k. 
Thus, no general effect can be easily formulated.

The text classification tasks seem more strongly affected by the batch size, up to 8\% according to the ALE plots, see Figure \ref{fig:ale_batch_size_sampling_rate}. 
While the sampling rates’ effect seems to be consistent across the datasets, the batch sizes' and sampling rates' effects differ between the model architectures. 
The overall positive effect for smaller batch sizes and sampling rates can be mostly attributed to the LSTM model, as the RNN model is not strongly affected by the batch size or sampling rate.
This implies, that the batch sizes' effect is not attributable to DP-SGD in general, but differs between scenarios. 
Further, this clearly highlights the importance of contextualizing conjectured hyperparameter effects and that it is necessary to test such conjectures for generalization.

To summarize, based on our results, the batch size does neither exhibit a strong main effect (as it explains only little variance and the amplitude of the ALE plots is mostly very small) nor exhibits a consistent effect across datasets or model architectures (as for some scenarios, the effect increases with increased batch sizes, for others the effect decreases, and for some the batch size does not seem to exhibit any effect at all). 
Thus, we \textbf{cannot replicate} results that suggest the conjectured effects of the batch size in DP-SGD, especially \emph{C1} that \emph{the batch size is the most important hyperparameter} to tune or \emph{C2} that \emph{increasing the batch size will increase model accuracy}.
These effects might hold for certain scenarios, but we cannot attribute a general batch size effect to DP-SGD.

\subsubsection{Number of Epochs}
As indicated by the fANOVA, the targeted number of epochs overall only has a small to negligible effect on the expected model accuracy across all sets.

\begin{figure*}[!h]
     \centering
     \subfloat{\includegraphics[width=0.5\textwidth]{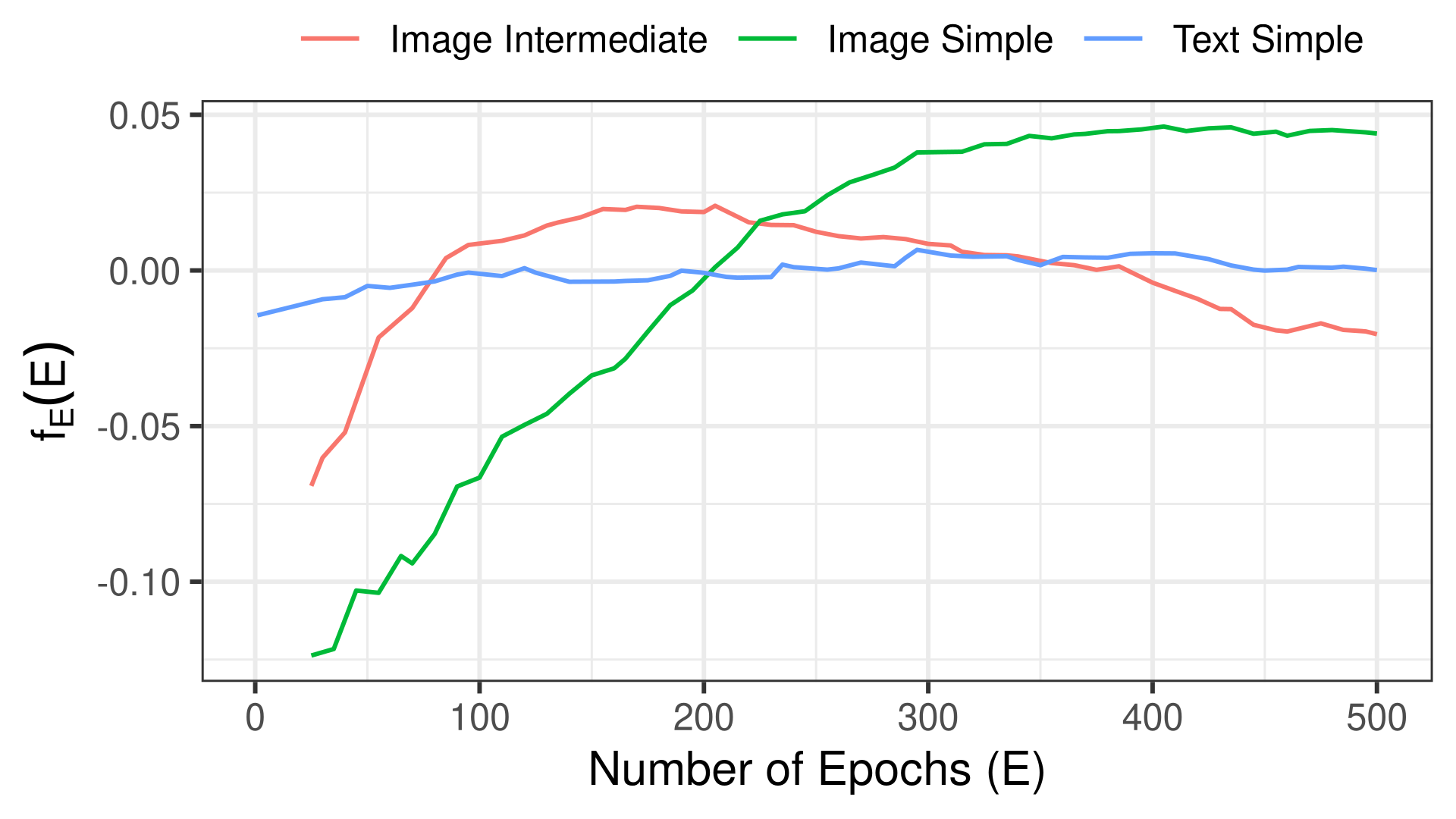}}
     \subfloat{\includegraphics[width=0.5\textwidth]{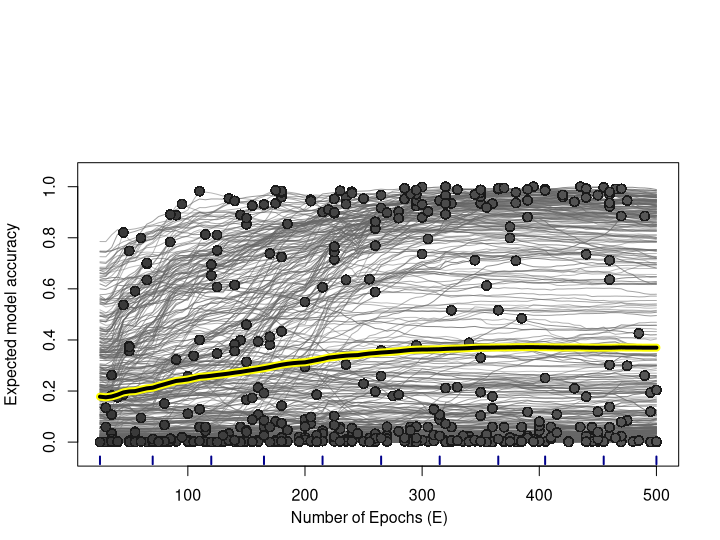}}
     \caption{ALE-plots for number of epochs across thee three experiment sets (left) and ICE plot for the number of epochs on the simple image task (right)}
    \label{fig:ale_epochs}
\end{figure*}

For the image classification tasks, the effects of the number of epochs is straightforward: In the simple task, the expected min-max normalized model accuracy increases by roughly 15\% up to 300 epochs and plateaus afterwards, while in the advanced image task, it increases by roughly 9\% up to its peak at 175 epochs and slightly decreases afterwards, see Figure \ref{fig:ale_epochs}.
In the advanced image task, for the CIFAR-100 dataset the expected model accuracy drops by up to 12\% for small epochs ($E<100$) compared to the ImageNette dataset. 
Besides this, there is no significant difference in the accumulated local effects across the levels of the categorical variables for the image classification tasks.
The positive effect is especially noticeable when not considering the averaged effect of the accumulated local effects or partial dependence plots, but the subset of ICE curves that predict a high accuracy, see Figures~\ref{fig:ale_epochs} and~\ref{fig:ice}.
The majority of curves in this subset increase in value first and then remain constant. 

The effect for the text classification task seems to be less strong according to the ALE plot.
In total, the number of epochs only affects the min-max normalized expected model accuracy by a maximum of~2\%, see Figure \ref{fig:ale_epochs}. 
However, similar to the batch size, the effect is not consistent across the model architectures, as the LSTM seems to be more affected by number of epochs (up to 4\% of expected model accuracy) than the RNN.
While this difference is clear, the overall amplitude of the effect is still small for the text classification task.

To summarize, based on our results, increasing the number of epochs (up to some threshold) does seem to increase the expected model accuracy for some tasks (i.e. image classification). 
However, for other tasks (i.e. text classification) the main effect seems to be negligible. 
We can \textbf{partially replicate} results suggesting \emph{C4} that \emph{the number of epochs has a significant effect on model accuracy}, in fact a positive effect, but only for the image classification task and not the text classification task.  

\subsubsection{Learning rate and clipping threshold}
\begin{figure*}[!h]
     \centering
     \subfloat{\includegraphics[width=0.5\textwidth]{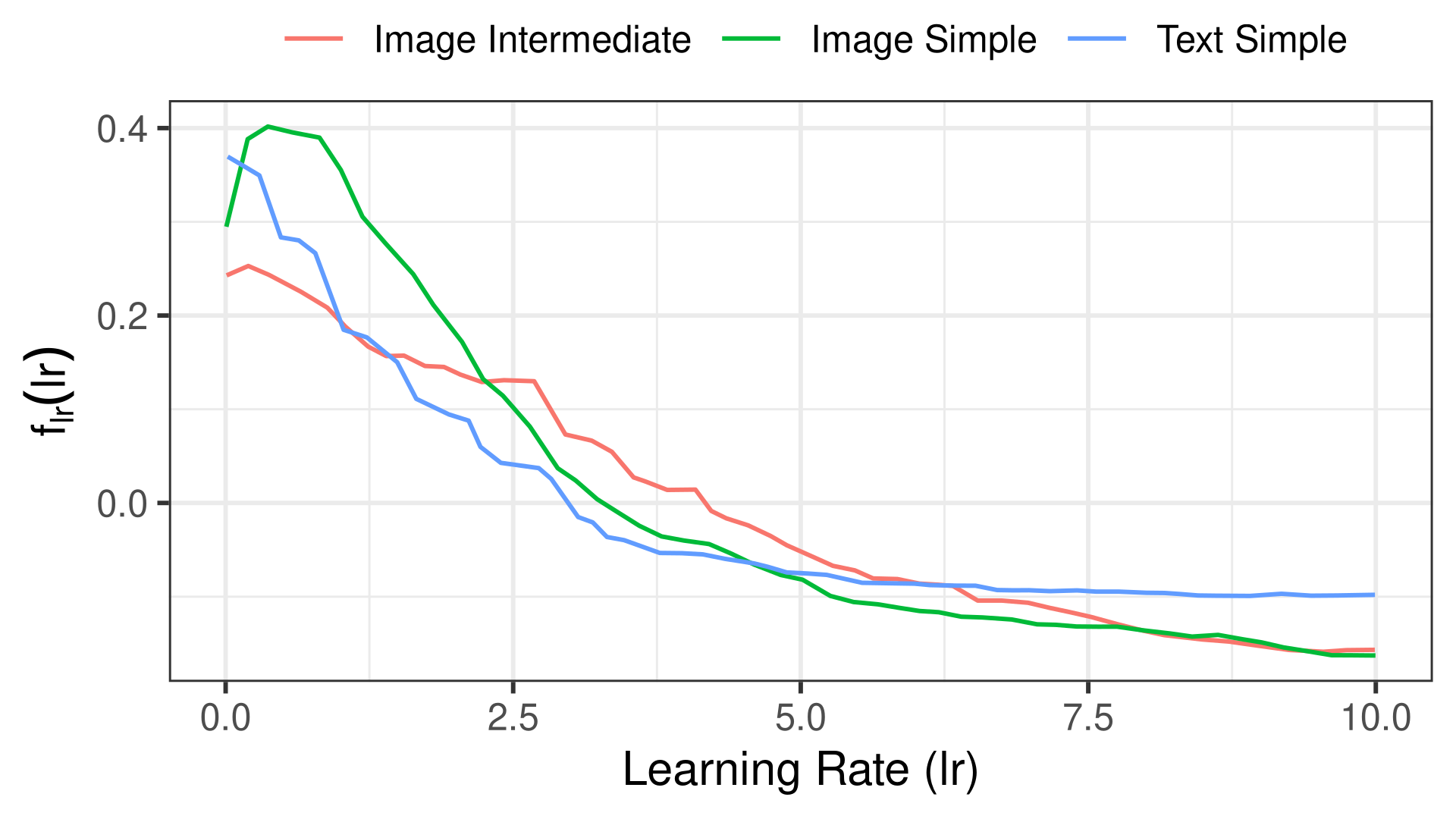}}
     \subfloat{\includegraphics[width=0.5\textwidth]{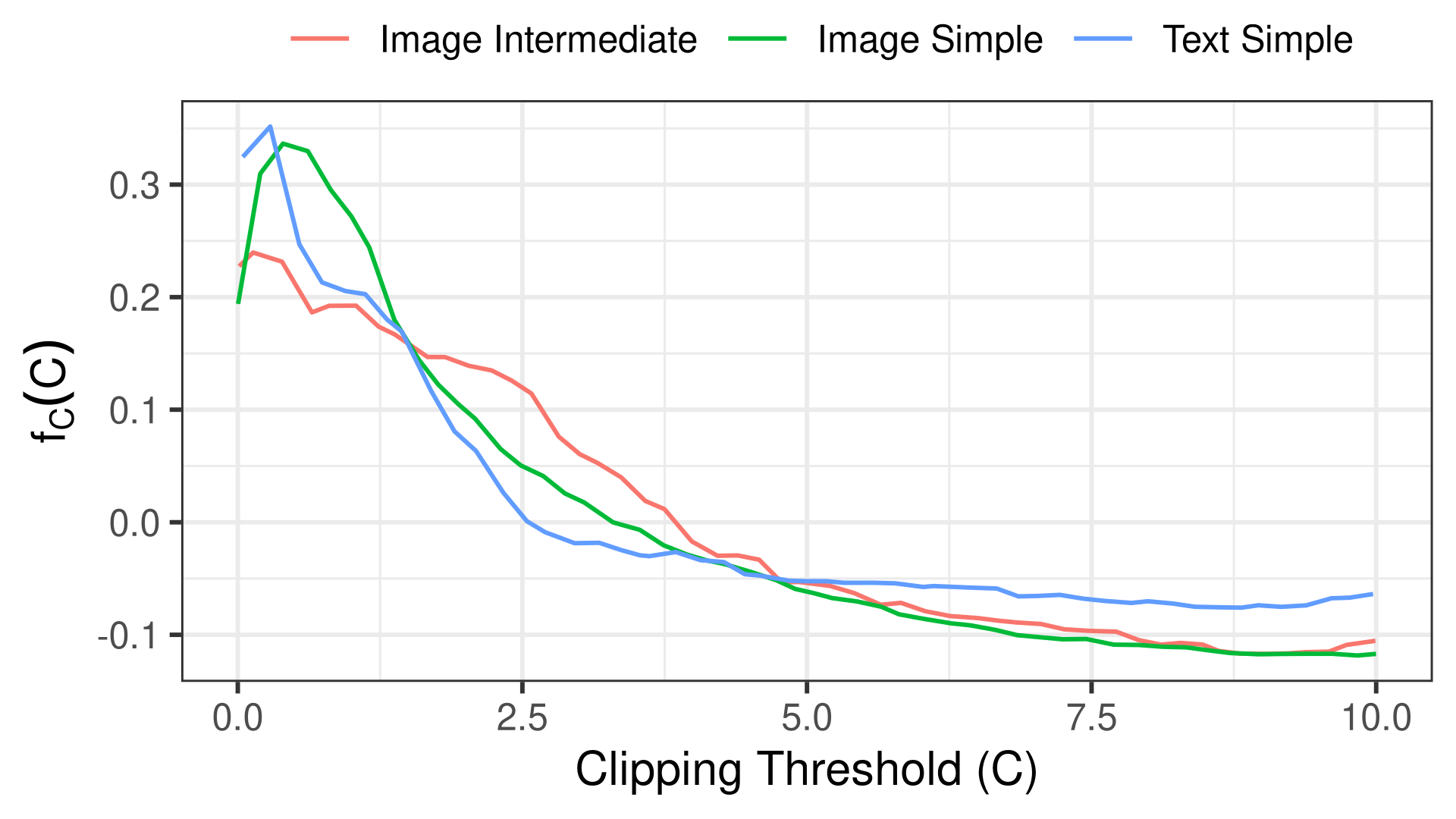}}
     \caption{ALE-plots for learning rate (left) and clipping threshold (right) across the three experiment sets}
    \label{fig:ale_lr_c}
\end{figure*}

As predicted by the fANOVA, the learning rate and clipping threshold have the strongest main effects, as they impact the expected min-max normalized model accuracy by up to 50\% according to the ALE plots, see Figure \ref{fig:ale_lr_c}. 
The effects of learning rate and clipping threshold seem to mostly mirror each other. 
The expected model accuracy is the highest for small values, peaks between $0.5$ and $1$, and then almost continuously decreases with increasing values.
However, for the intermediate image classification task, the effect decreases considerably slower when compared to the simple image or text classification tasks. 
Their effects do not differ significantly between the sets or across the levels of the categorical values inside each set.

\subsection{Interaction Effects}
Informed by the fANOVA, we investigate the most promising interaction effects in descending order. 
For brevity, we only show plots for the simple image classification task, except when specifically mentioned otherwise.

\subsubsection{Learning rate and clipping threshold}
\begin{figure*}[]
     \centering
     \subfloat{\includegraphics[width=0.5\textwidth]{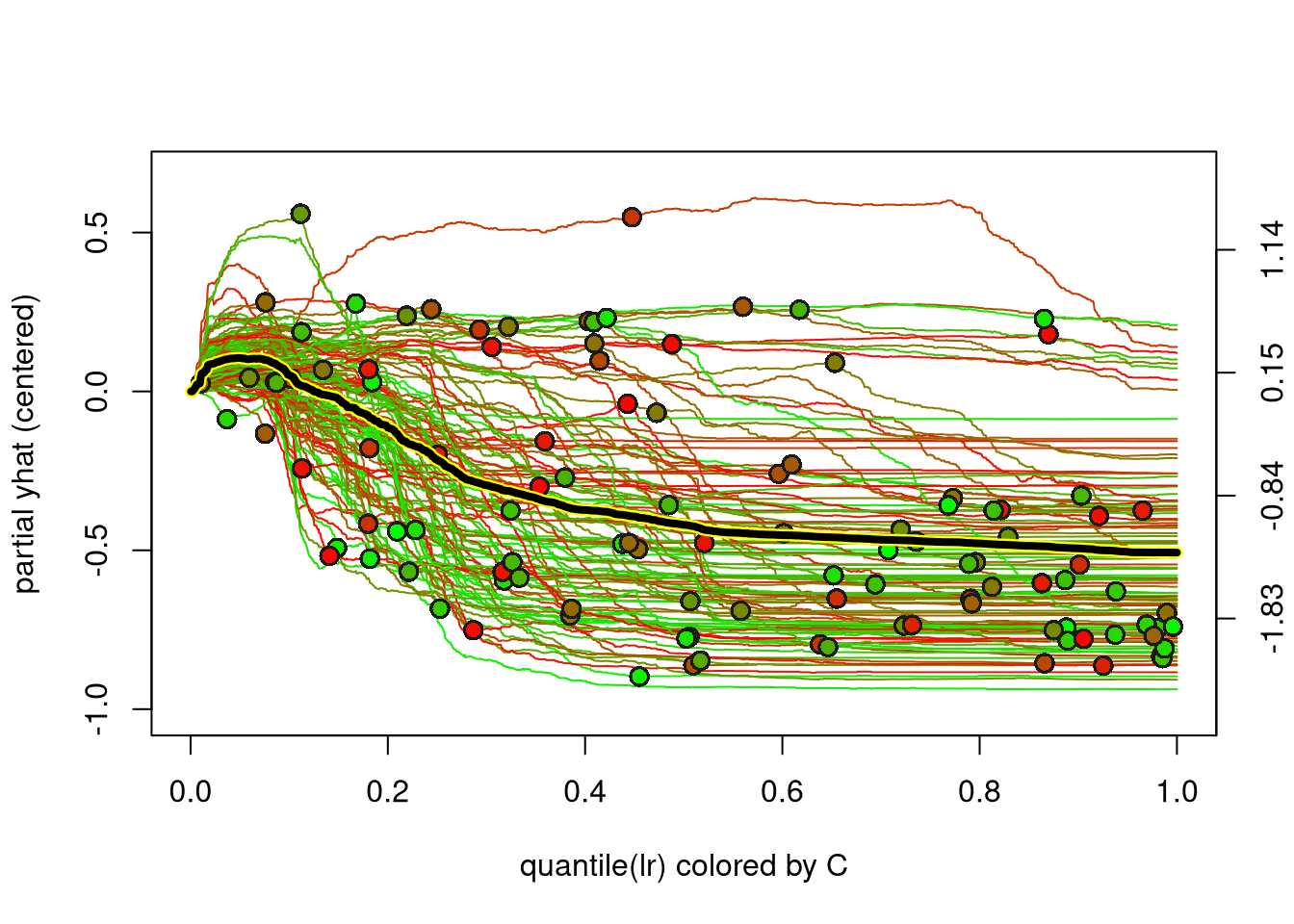}}
     \subfloat{\includegraphics[width=0.5\textwidth]{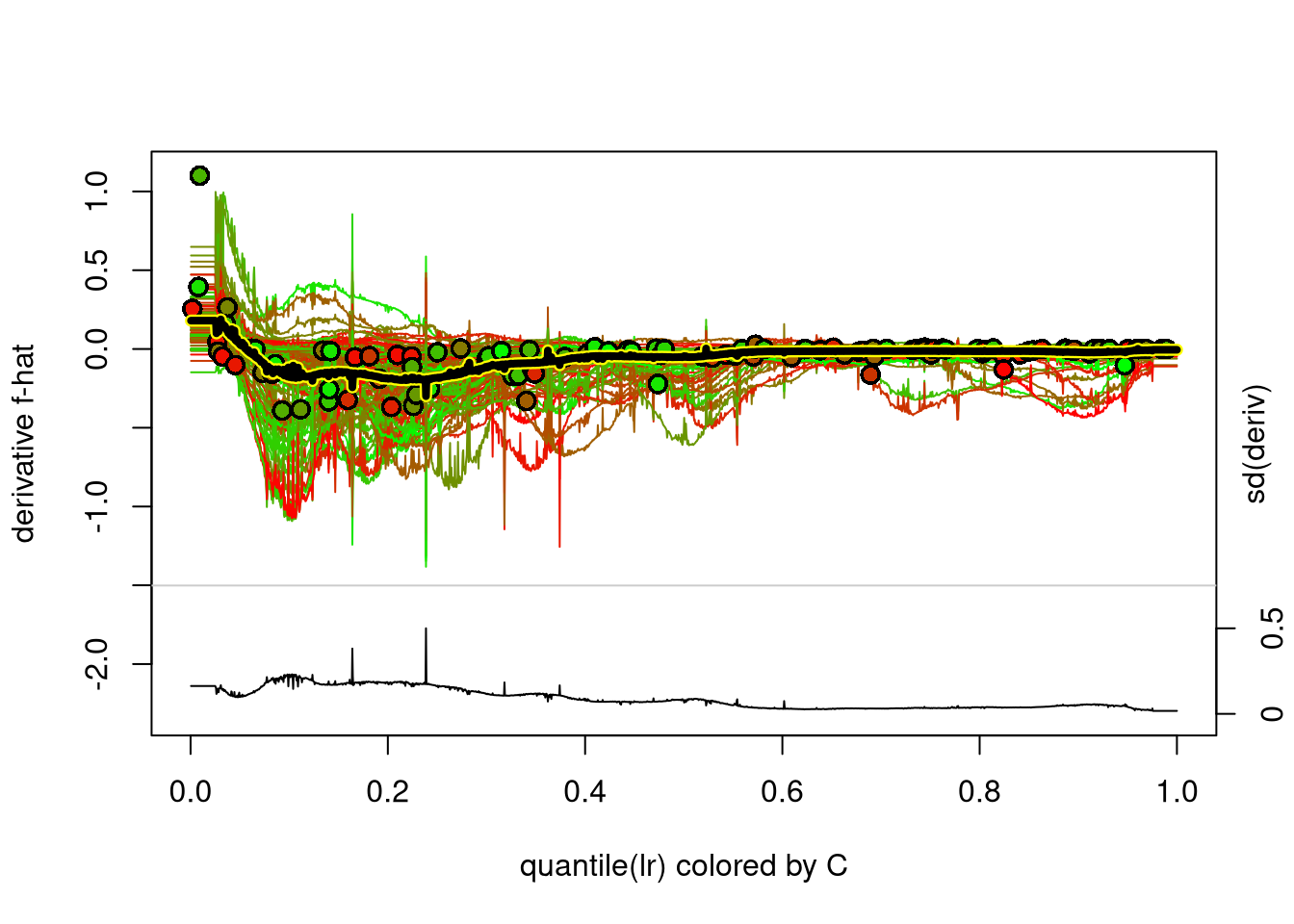}}
     \caption{Colored centered ICE (left) and derivative ICE (right) plots for the learning rate ($lr$) colored by the clipping threshold ($C$) for the simple image classification task. Red lines corresponds to low, green to high values of the clipping threshold ($C$).}
     \label{fig:c_d_ice_lr_by_C}
\end{figure*}

To recapitulate, the condensed main effect on the expected accuracy is that by increasing the clipping threshold or learning rate, the expected accuracy decreases. However, as can be seen from the centered ICE plots (see Figure \ref{fig:c_d_ice_lr_by_C}), the individual lines diverge from the average line.
The effect of the variable is thus not consistent across the range of the other variables but rather indicates interaction effects with other variables. 
If the effect of a variable would be independent of the other variables, we would expect all the lines in the centered ICE plot to be on top of each other and the same as the average line.

The d-ICE plots (see Figure \ref{fig:c_d_ice_lr_by_C} and \ref{fig:c_d_ice_C_by_lr}) further showcase that the variance of the derivatives of the individual curves is not consistent across the learning rate's (or clipping threshold's) range. 
This indicates that the interaction effect on the learning rate is not consistent across the learning rate's range, but rather that the learning rate's effect is affected differentially by another variable across the learning rate's range (and vice versa for the clipping threshold). 
The interaction seems to be stronger for lower learning rate values (or clipping threshold) than for higher ones.
Or in other words, varying the clipping threshold will affect the expected model accuracy more if the learning rate is small compared to if it is large.

As indicated by the fANOVA, the interaction in question is between the learning rate and clipping threshold. Thus, to analyze the nature of this interaction, we colored the ICE plots of one variable with the value of the other. Red corresponds to low values, green to high values (see Figure \ref{fig:c_d_ice_lr_by_C}, \ref{fig:c-ice_2d-ale_epochs_lr}, and \ref{fig:c_d_ice_C_by_lr}).
Unfortunately, the interactions between the variables are not clearly deducible from this visualization, as no clear or obvious pattern emerges. 
At most, one could hypothesize/speculate trends, such that high values (green) tend to have more extreme negative derivatives, which would indicate that high values have a stronger effect than lower ones.

\begin{figure*}
     \centering
     \subfloat{\includegraphics[width=0.32\textwidth]{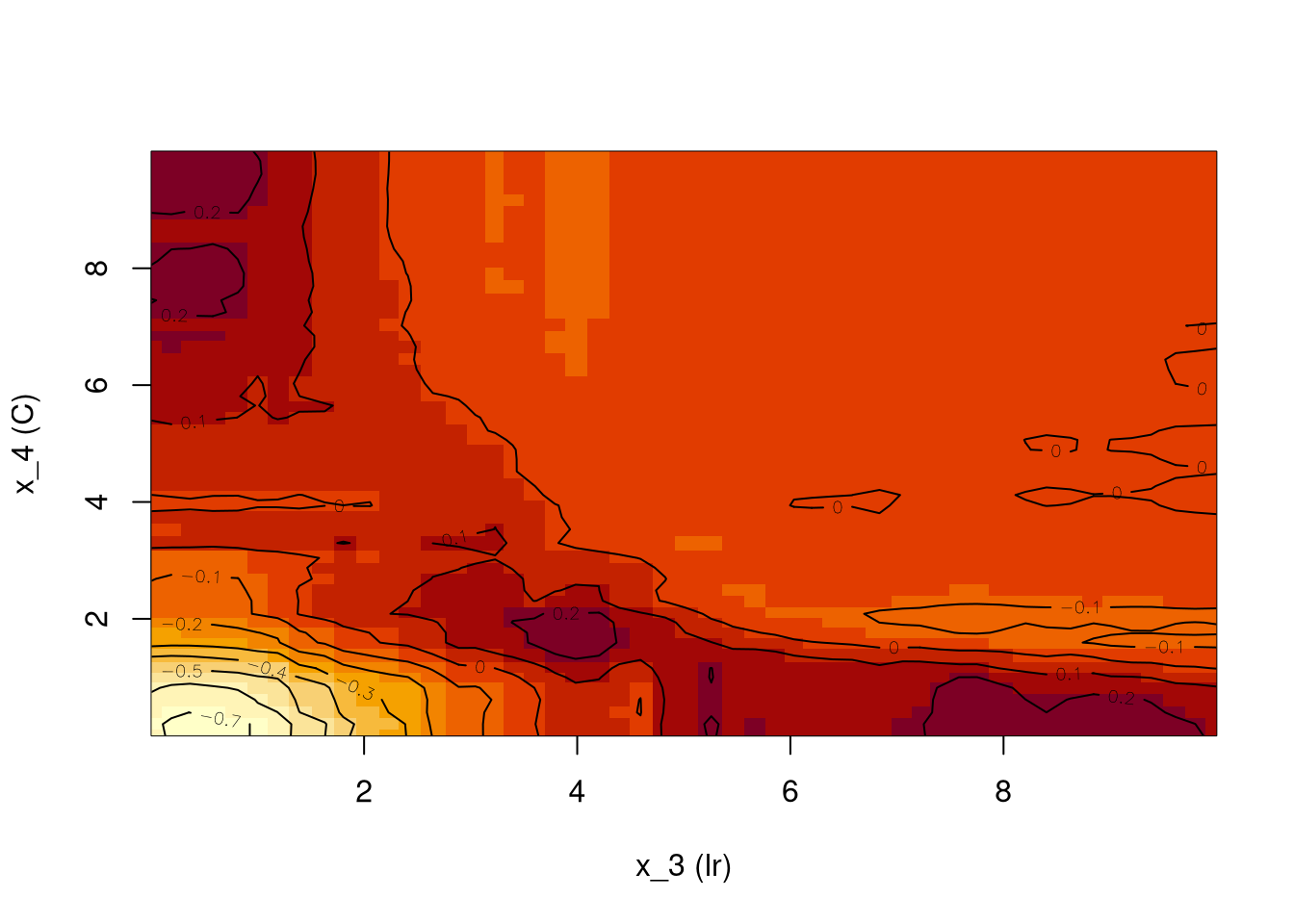}}
     \subfloat{\includegraphics[width=0.32\textwidth]{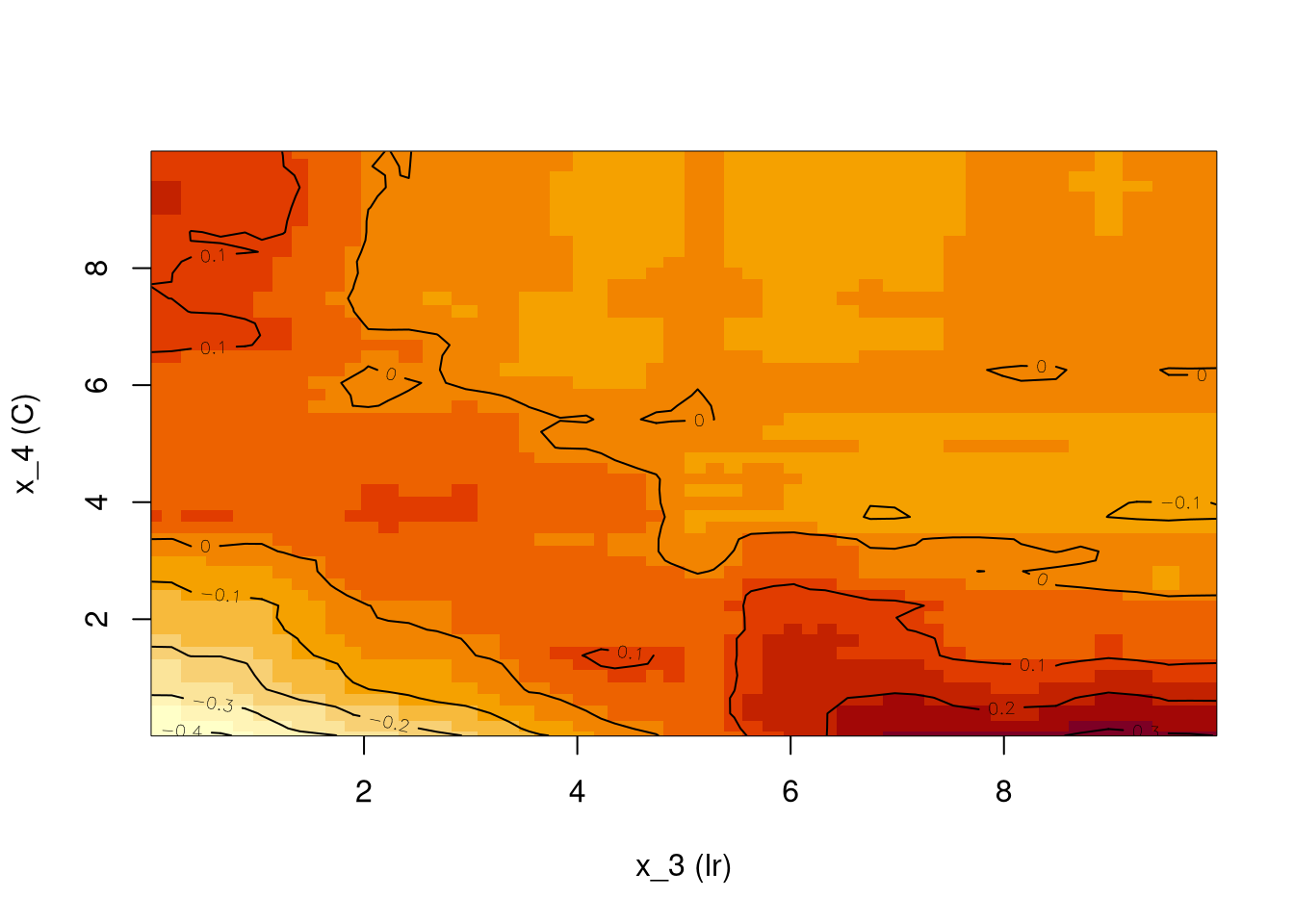}}
     \subfloat{\includegraphics[width=0.32\textwidth]{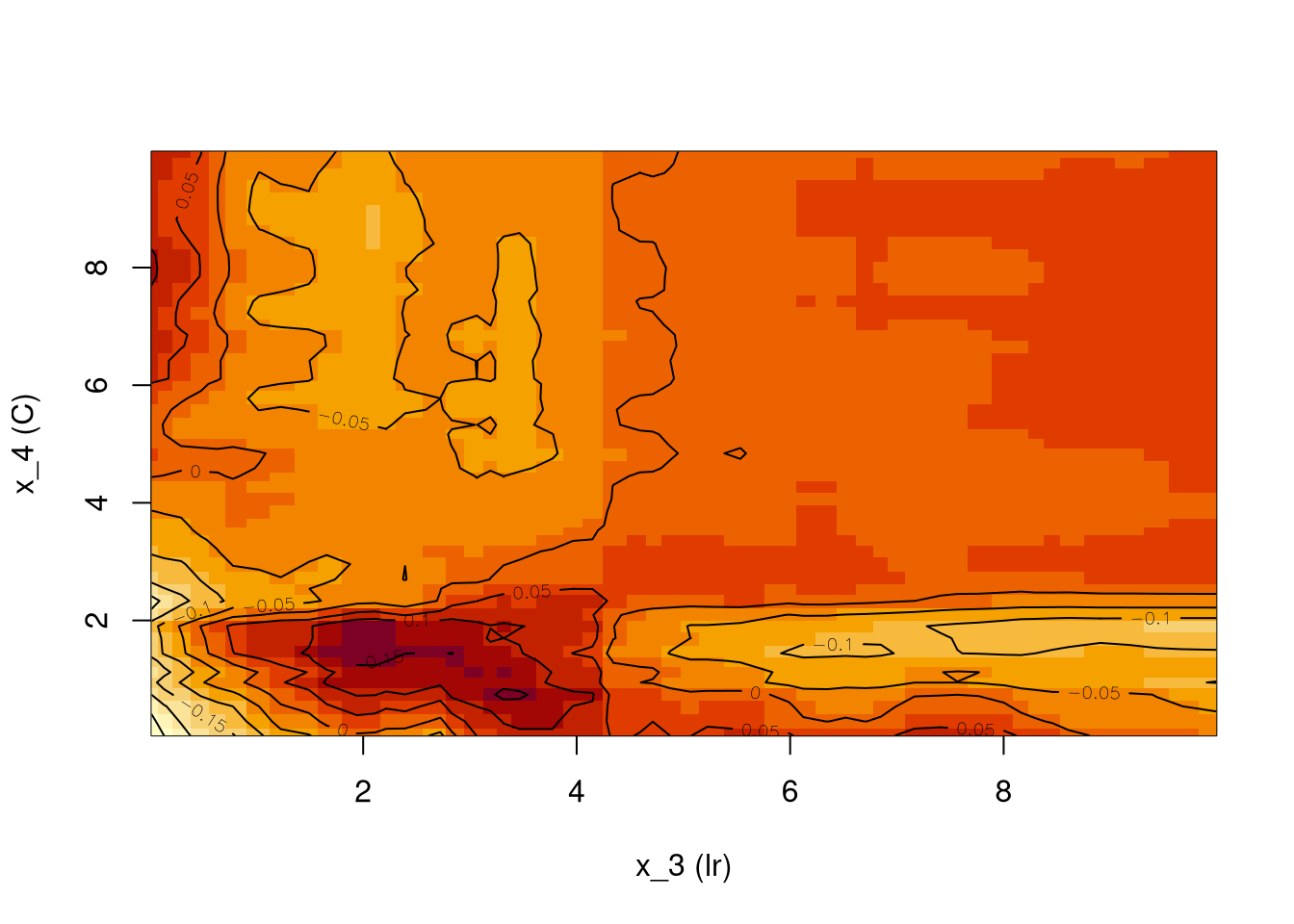}}
     \caption{Two dimensional ALE-plots for learning rate ($x$-axis) and clipping threshold ($y$-axis) on the simple image task (left), intermediate image task (center), and the simple text task (left). Darker colors represent high, lighter colors represent low expected model accuracy values.}
    \label{fig:ale-2d_lr_c}
\end{figure*}

To test for trends in the interaction we additionally use ALE interaction plots, here for all three sets (simple image, advanced image and simple text), see Figure \ref{fig:ale-2d_lr_c}.
The plots for the image classification tasks clearly show a relationship between the learning rate and the clipping threshold (even stronger for the simple tasks). 
While the main effects do suggest that small values between $0.5$ and $1$ yield the highest expected performance for both hyperparameters, the interaction plot shows that fixing both variables at the same time to this value would lead to bad expected performance. 
Instead, it seems optimal to pair large values of one hyperparameter with small values of the other or to pair medium values together. 
This interaction effect seems to be less strong for the advanced image classification task than for the simple one, but still clearly observable.

Interestingly, while the main effects of both, learning rate and clipping threshold, are similar between the image and text classification tasks, the interaction effect seems to be slightly different. 
While very low values of one hyperparameter can be paired with high values of the other to gain a higher expected model accuracy than average, the area seems to be significantly less pronounced when compared to the image classification tasks. 
Furthermore, combinations that would lead to a higher than average expected model accuracy for the image classifications tasks would be below average expected model accuracy on the text classification task. 
Instead, and in contrast to image classification, the most promising area for higher than average performance seems to be with combination of small (but not tiny) values.

To summarize, based on our results, the learning rate ($lr$) and clipping threshold ($C$) exhibit strong and consistent main effects.
However, more importantly, the effect of the learning rate heavily depends on the clipping threshold, and vice versa.
Considering both in isolation may lead to unnecessary low expected model accuracy, as the effect of one hyperparameter is significantly altered by the value of the other. 
Thus, these two hyperparameters have to be considered in conjunction.

For the image classification tasks, our results clearly replicate findings from prior works that conjectured there to be an interaction effect, that is, there is a constant such that  hyperparameters that lie on a $lr \cdot C = \textrm{const}$ curve exhibit a higher expected model accuracy than hyperparameter tuples that do not lie on this curve. 
For the text classification tasks, our results only partially exhibit this effect, weakening this hypothesis. 
However, we can \textbf{replicate} results that support \emph{C5} that \emph{the clipping threshold's effect is affected by the learning rate} and  \emph{C6} that \emph{the learning rate's effect is affected by the clipping threshold}.
Due to this interaction effect, the effect of the learning rate heavily depends on the value of the clipping threshold and vice versa. Isolated conjectures from related work about the effect of the learning rate or clipping threshold therefore seem to be ill-informed. 
This further highlights the importance of proper factorial experiment design to test for interactions instead of an one-factor-at-a-time experiment.

\subsubsection{Epochs and learning rate or clipping threshold}
\begin{figure*}[!ht]
     \centering
     \subfloat{\includegraphics[width=0.5\textwidth]{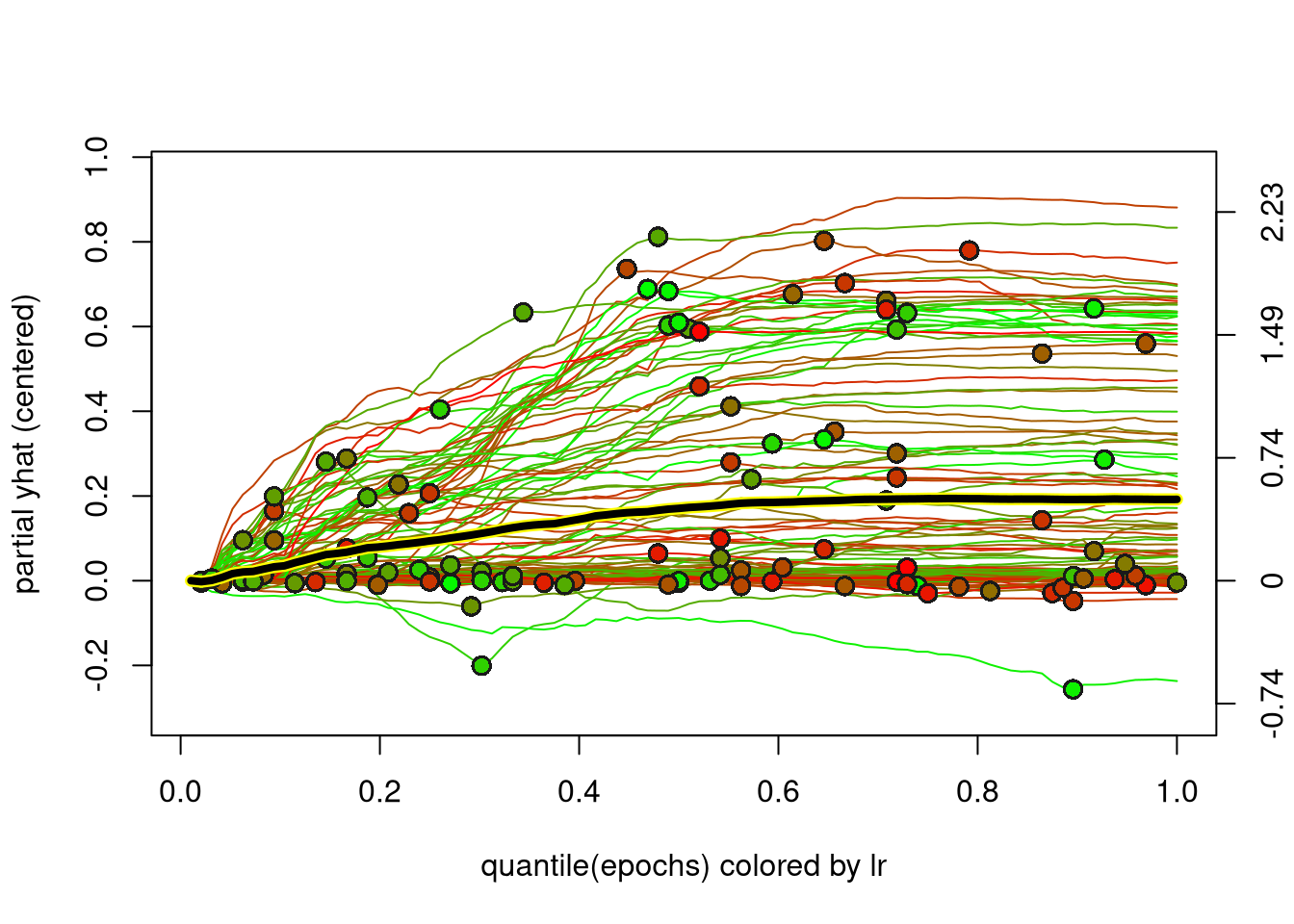}}
     \subfloat{\includegraphics[width=0.5\textwidth]{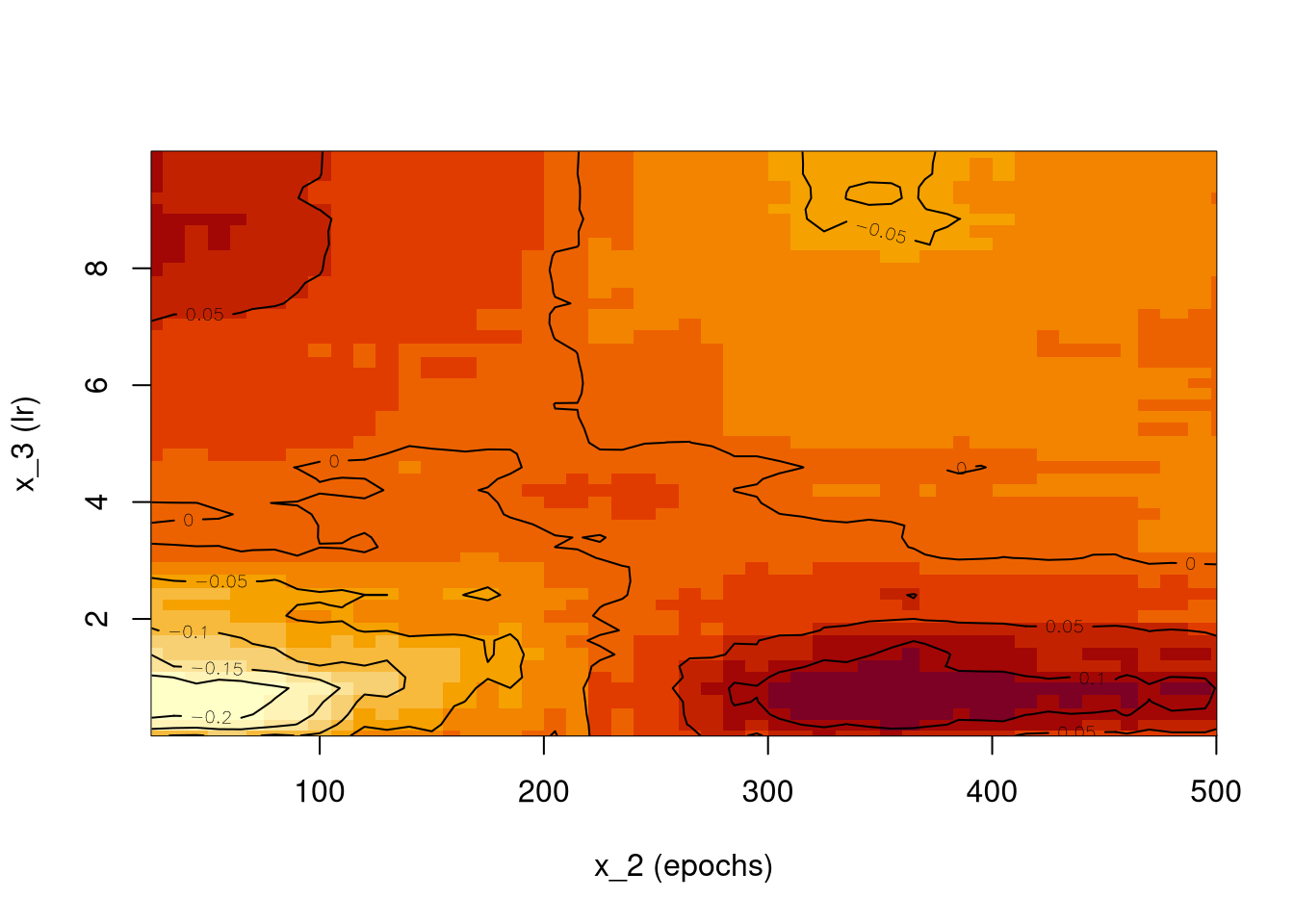}}
     \caption{Colored centered ICE plot (left) for the number of epochs ($E$) colored by the learning rate ($lr$) for the simple image classification task. Red lines corresponds to low, green to high values of the learning rate ($lr$). Two dimensional ALE plot (right) for the number of epochs ($x$-axis) and learning rate ($y$-axis) for the simple image classification task. Darker colors represent high, lighter colors represent low expected model accuracy values.}
     \label{fig:c-ice_2d-ale_epochs_lr}
\end{figure*}

The other possibly interesting interaction effect according to the fANOVA is between the number of epochs ($E$) and learning rate ($lr$) or clipping threshold ($C$). 
Figure \ref{fig:c-ice_2d-ale_epochs_lr} shows the ICE and two dimensional ALE interaction plots for the interaction with the learning on the simple image tasks (see the Appendix for the plot describing the interaction with the clipping threshold).

Similarly as with the learning rate or clipping threshold, while the ICE plots provide a clear indication of the existence of interaction effects, the nature of the interaction is hard to grasp. 
However, the ALE plots indicate a clear but not necessarily very strong interaction between learning rate (or clipping threshold) and the number of epochs on the simple image classification task. 
The effect of few epochs seem to be negatively impacted by small learning rates (or clipping thresholds) and positively by large ones. 
The positive effect of a large number of epochs seem to be amplified by small learning rates (or clipping thresholds) and reduced by large ones.
Even though this effect is negligible for the other sets in our experiment, this interaction effect has not been observed by the related work so far, warranting further analysis in future work. 
\subsubsection{Batch size and number of epochs}
Even though the fANOVA already indicates that the interaction of the number of epochs ($E$) and the batch size ($B$) is not strong, discussing this interaction effect is still of interest.
Please recall that the combination of both determines the noise multiplier ($\sigma$) -- and thus, in combination with the clipping threshold ($C$), the amount of Gaussian noise that is injected during the training ($Z\sim\mathcal{N}(0,\sigma^2C^2I)$).
Related work conjectured that to avoid a large noise multiplier, one should balance large batch sizes with fewer epochs.
This is based on the implicit assumption that the accuracy loss of DP-SGD is primarily attributable to the added noise, and thus, that reducing the amount of noise added is the most promising avenue for closing the accuracy gap between SGD and DP-SGD.
Thus, if large noise multipliers would be generally detrimental to model accuracy, there would be an interaction effect such that large batch sizes and high number of epochs would decrease the expected model accuracy.
However, as we cannot identify an interaction effect between the number of epochs and batch size, we cannot confirm this conjecture.
Instead, it appears that in some scenarios it even is beneficial to choose a large number of epochs, even though this implies a large noise multiplier, which intuitively seems disadvantageous. 
To summarize, based on our result, we \textbf{cannot replicate} results that suggest \emph{C3} that \emph{the batch size has interaction effects} and especially not that the batch size has a negative interaction effect with the number of epochs.

\section{Conclusions}
\label{sec:conclusion}
Investigating the effects of hyperparameters in DP-SGD is an interesting endeavor, not only for practice, but also for research in order to better understand the privacy-utility trade-off in differentially private machine learning. 
In this work, we tried to replicate conjectures from prior research about the effects of hyperparameters when training machine learning models with DP-SGD.
By conducting the largest independent and dedicated experiment and subsequently identifying main and interaction effects, we were able to replicate the strong and important relationship between the clipping threshold and the learning rate.
However, we were not able to consistently (across datasets, model architectures, and differential privacy budgets) replicate other conjectures, such as an interaction between the number of epochs and the batch size.

Our results imply, especially for practice, that one cannot circumvent hyperparameter optimization entirely. 
However, even though there is no free lunch when it comes to hyperparameter configurations in machine learning \cite{wolpert_lack_1996}, insights on the hyperparameter effects can still be valuable in practice to speed up hyperparameter optimizations. 
For example, the inverse relationship between learning rate and clipping threshold can serve as a solid prior for warm-starting model based hyperparameter optimization methods, as similar learning tasks do correlate with similar hyperparameter configurations \cite{giraud-carrier_toward_2005, feurer_hyperparameter_2019, vanschoren_meta-learning_2019}.
Furthermore, understanding hyperparameter effects of learning algorithms is not only beneficial for maximizing model accuracy, but also for practitioners to increase their trust into models when deploying them into production \cite{drozdal_trust_2020}.
This naturally extends to understanding the privacy-utility trade-off in DP-SGD. 
Using interpretable machine learning methods in general, and the methods we applied specifically, has been demonstrated to be useful to foster the understanding of hyperparameter effects in different contexts \cite{moosbauer_explaining_2021}.

As many insights on the hyperparameter effects from related works were drawn from experiments that were not primarily designed to identify such effects, it is not necessarily surprising that we were not able to replicate all of the conjectured effects. 
Rather, it highlights that the identification of effects of independent variables on dependent variables can be sensitive to the experiment design.
While some effects might be distinctly identifiable in certain scenarios, on certain datasets, or with certain architectures, its is necessary to distinguish whether these effects are attributable to the learning algorithm or the specific scenario. 
Thus, dedicated, independent and factorial replication studies like this one are refining the scientific body of knowledge.  

\section*{Acknowledgment}
Funded in part by the German Research Foundation (DFG, Deutsche Forschungsgemeinschaft) as part of Germany’s Excellence Strategy – EXC 2050/1 – Project ID 390696704 – Cluster of Excellence “Centre for Tactile Internet with Human-in-the-Loop” (CeTI) of Technische Universität Dresden and by the German Ministry of Education and Research (Project SynthiClick, FKZ 16KISA108K).
Further, this work was supported by KASTEL Security Research Labs, Karlsruhe.
We thank the inhouse textician at KASTEL Security Research Labs and  the anonymous reviewers for their feedback and valuable input.

\bibliographystyle{IEEEtran}
\bibliography{DPML}

\begin{figure*}
     \centering
     \begin{subfigure}[b]{0.49\textwidth}
         \centering
         \includegraphics[width=\textwidth]{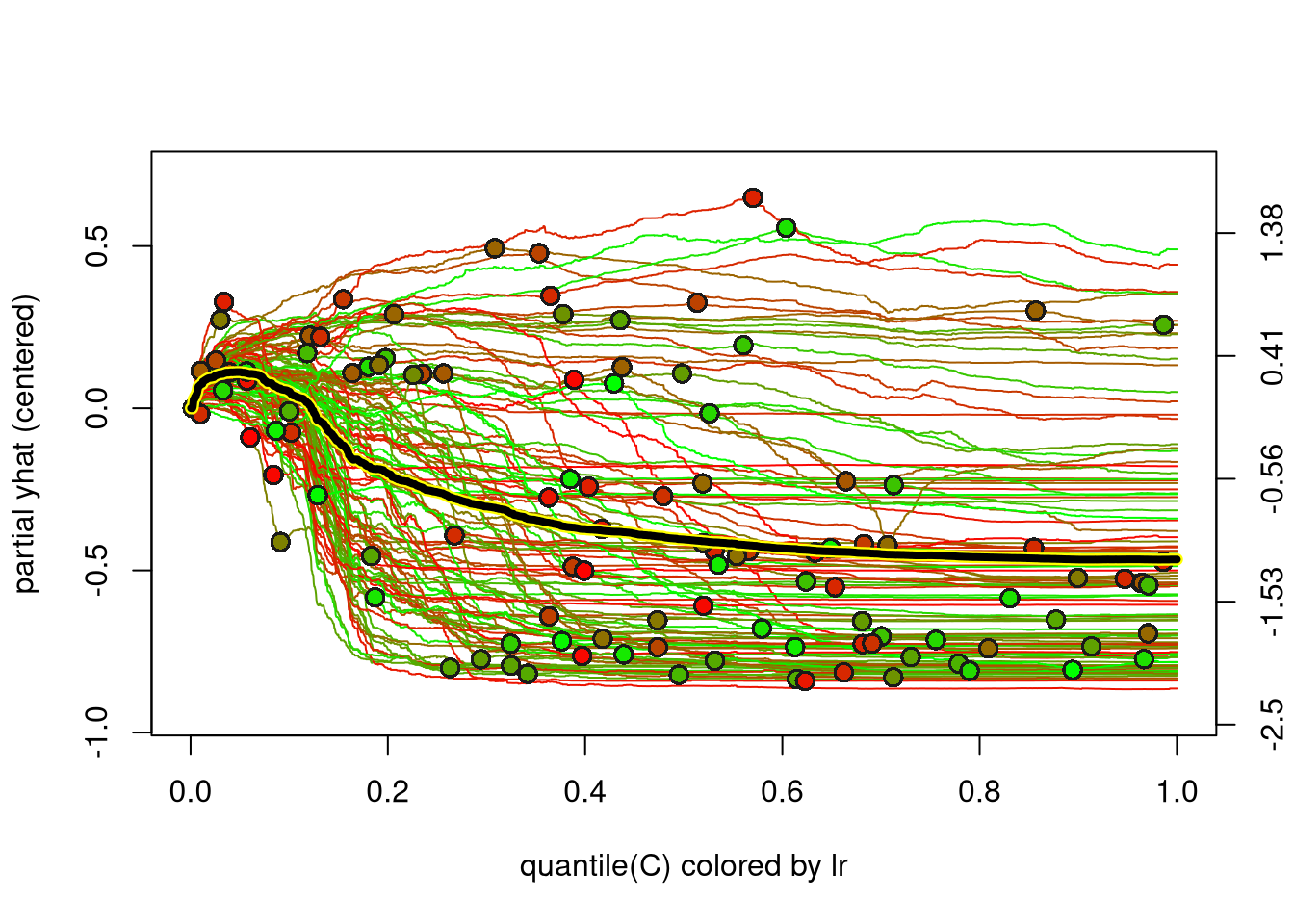}
     \end{subfigure}
     \hfill
     \begin{subfigure}[b]{0.49\textwidth}
         \centering
         \includegraphics[width=\textwidth]{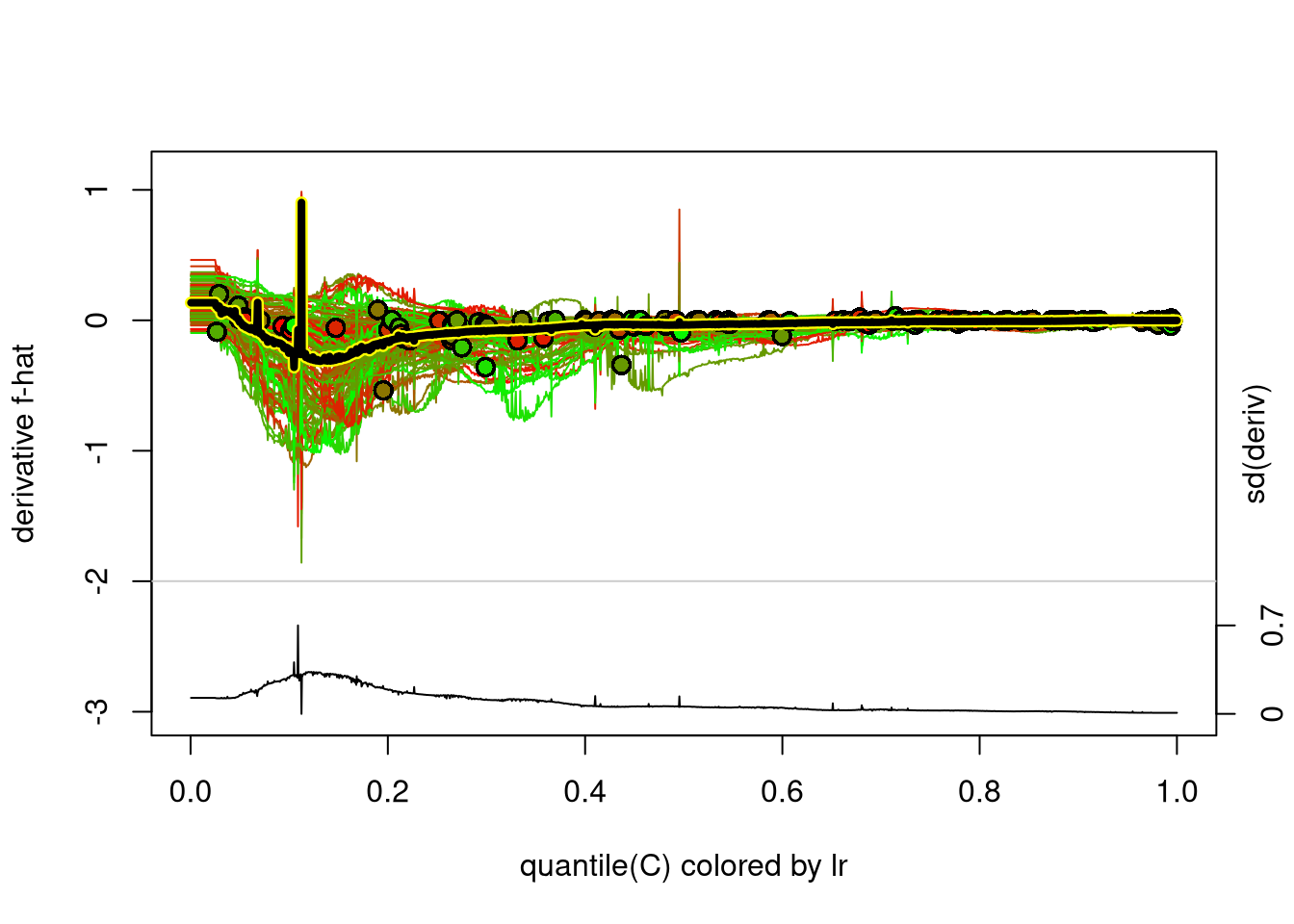}
     \end{subfigure}
     \caption{Colored centered ICE (left) and derivative ICE (right) plots for the clipping threshold ($C$) colored by the learning rate ($lr$) for the simple image classification task. Red lines corresponds to low, green to high values of the learning rate ($lr$).}
     \label{fig:c_d_ice_C_by_lr}
\end{figure*}

\begin{figure*}[!h]
     \centering
     \begin{subfigure}[b]{0.49\textwidth}
         \centering
        \includegraphics[width=\textwidth]{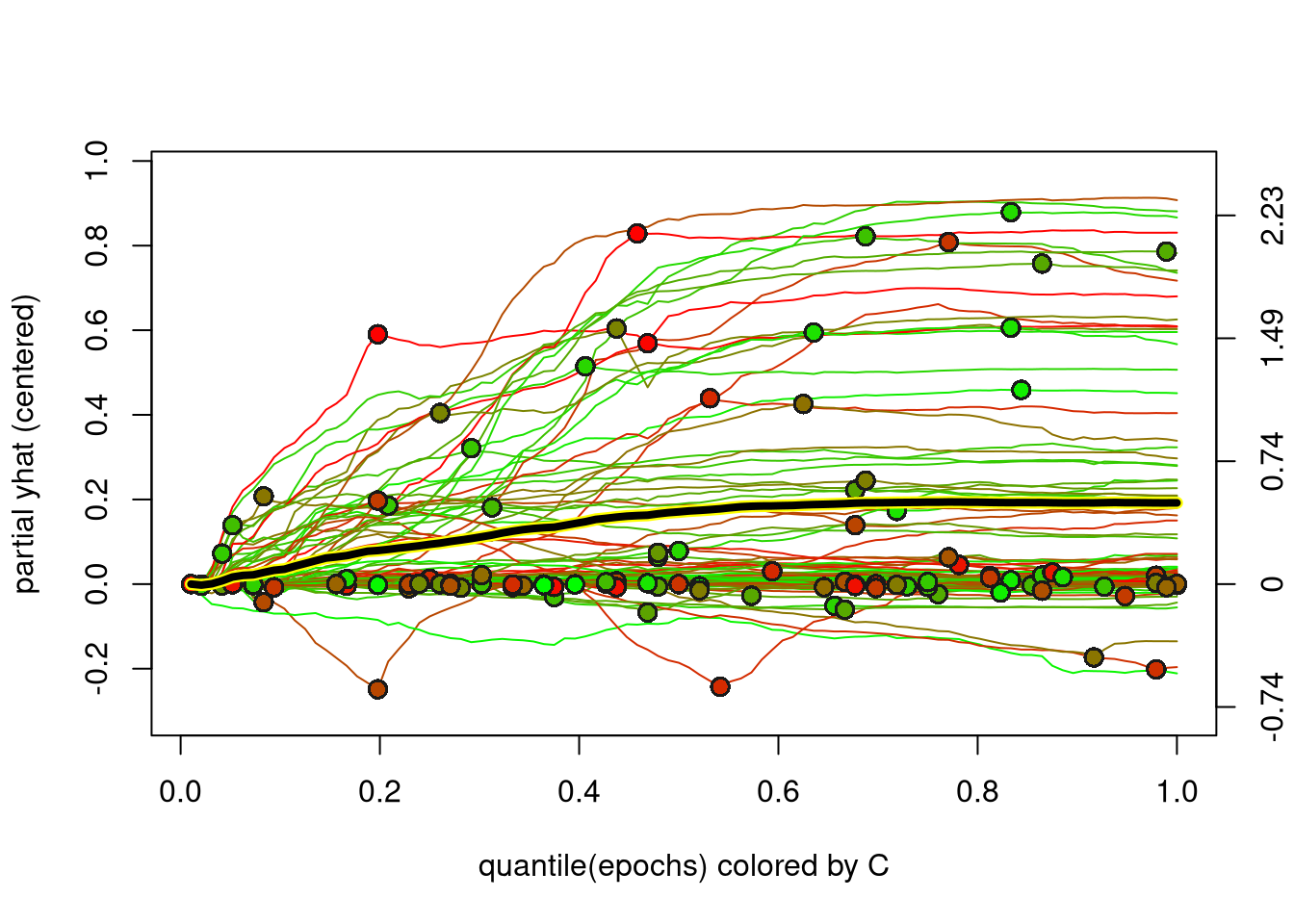}
     \end{subfigure}
     \hfill
     \begin{subfigure}[b]{0.49\textwidth}
         \centering
         \includegraphics[width=\textwidth]{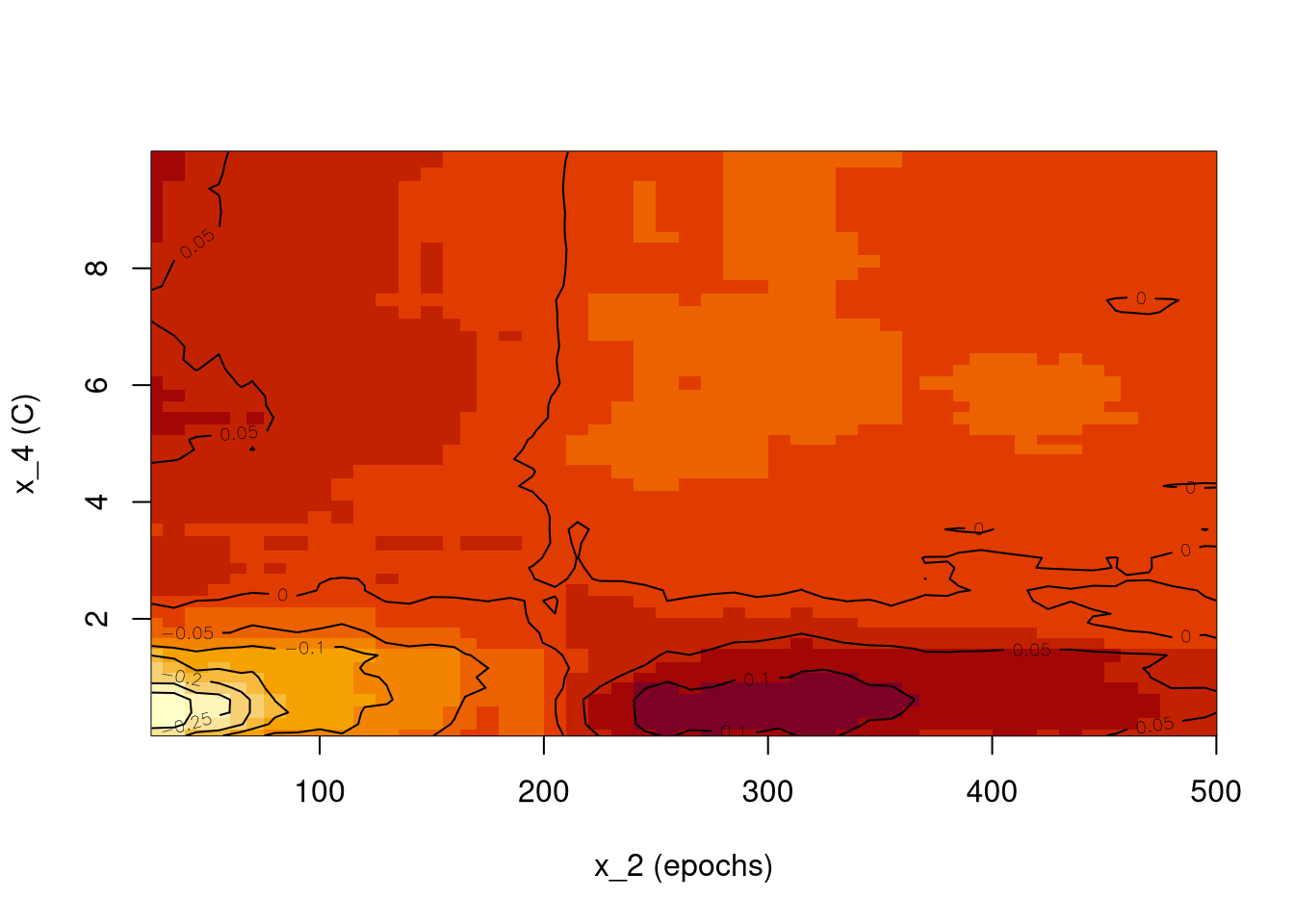}
     \end{subfigure}
     \caption{Colored centered ICE plot (left) for the number of epochs ($E$) colored by the clipping threshold ($C$) for the simple image classification task. Red lines corresponds to low, green to high values of the clipping threshold ($C$). Two dimensional ALE plot (right) for the number of epochs ($x$-axis) and clipping threshold ($y$-axis) for the simple image classification task. Darker colors represent high, lighter colors represent low expected model accuracy values.}
     \label{fig:c-ice_2d-ale_epochs_C}
\end{figure*}

\begin{figure*}[ht]
     \centering
     \begin{subfigure}[b]{0.32\textwidth}
         \centering
         \includegraphics[width=\textwidth]{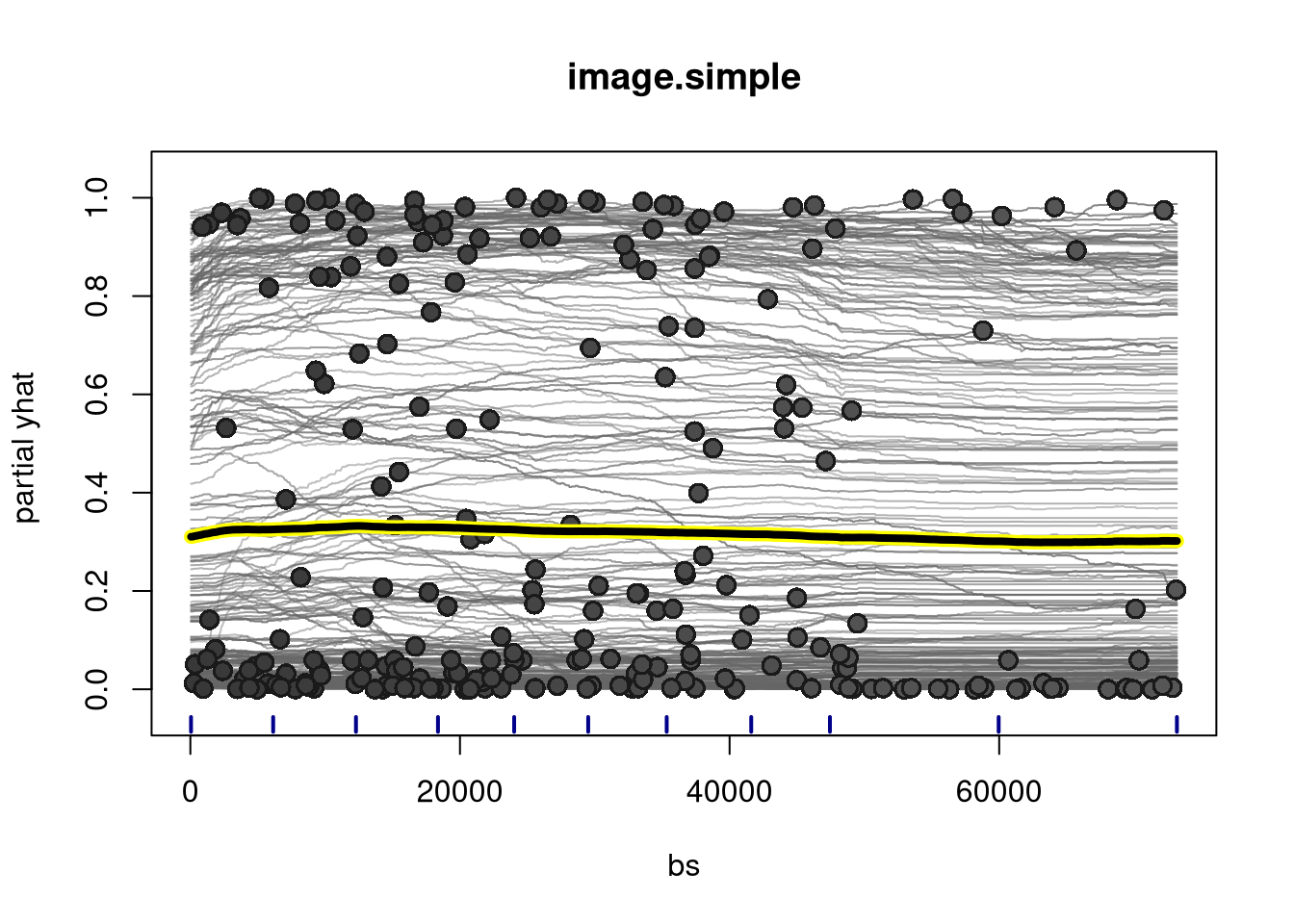}
     \end{subfigure}
     \hfill
     \begin{subfigure}[b]{0.32\textwidth}
        \centering
     \includegraphics[width=\textwidth]{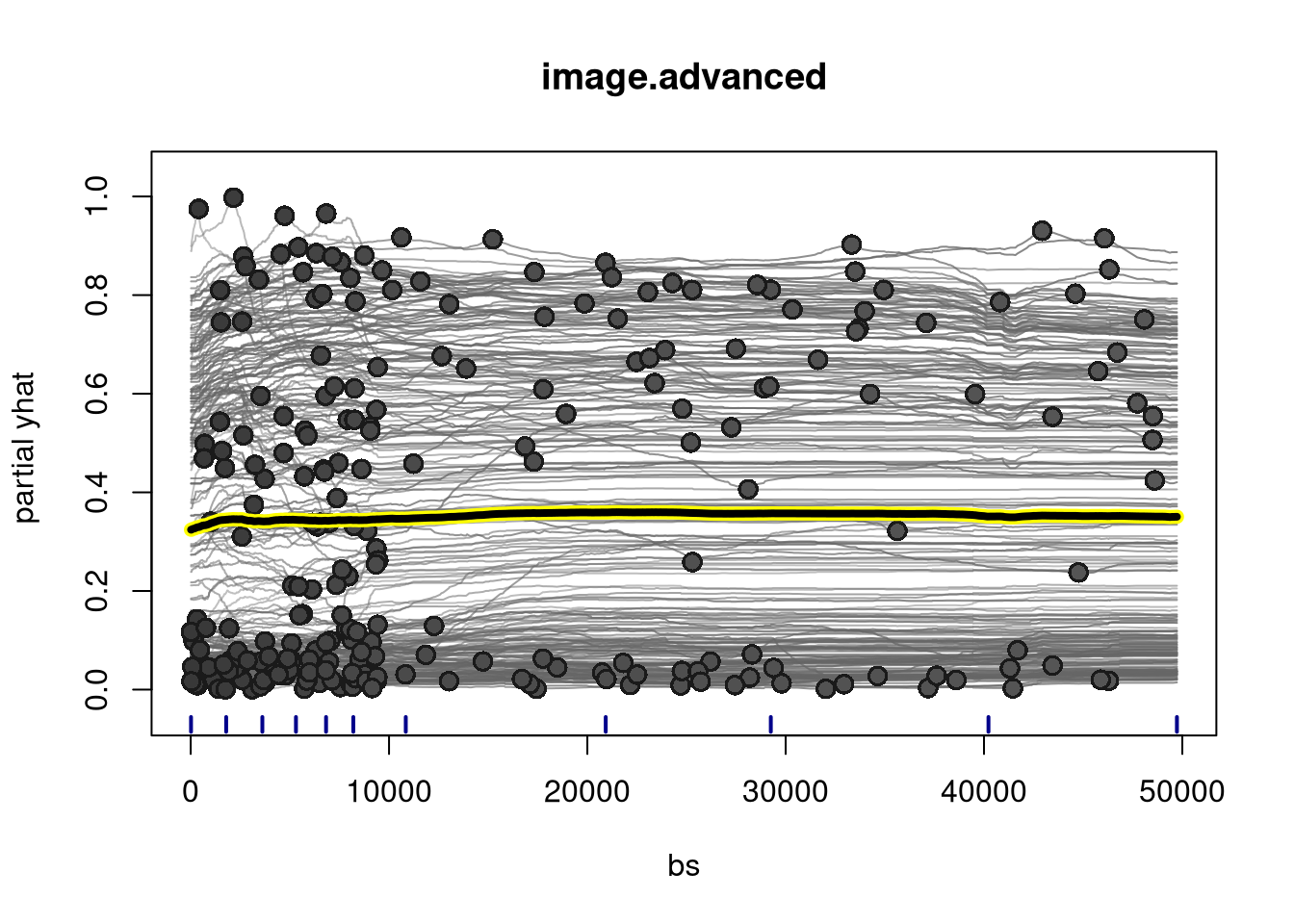}
     \end{subfigure}
     \hfill
     \begin{subfigure}[b]{0.32\textwidth}
     \centering
     \includegraphics[width=\textwidth]{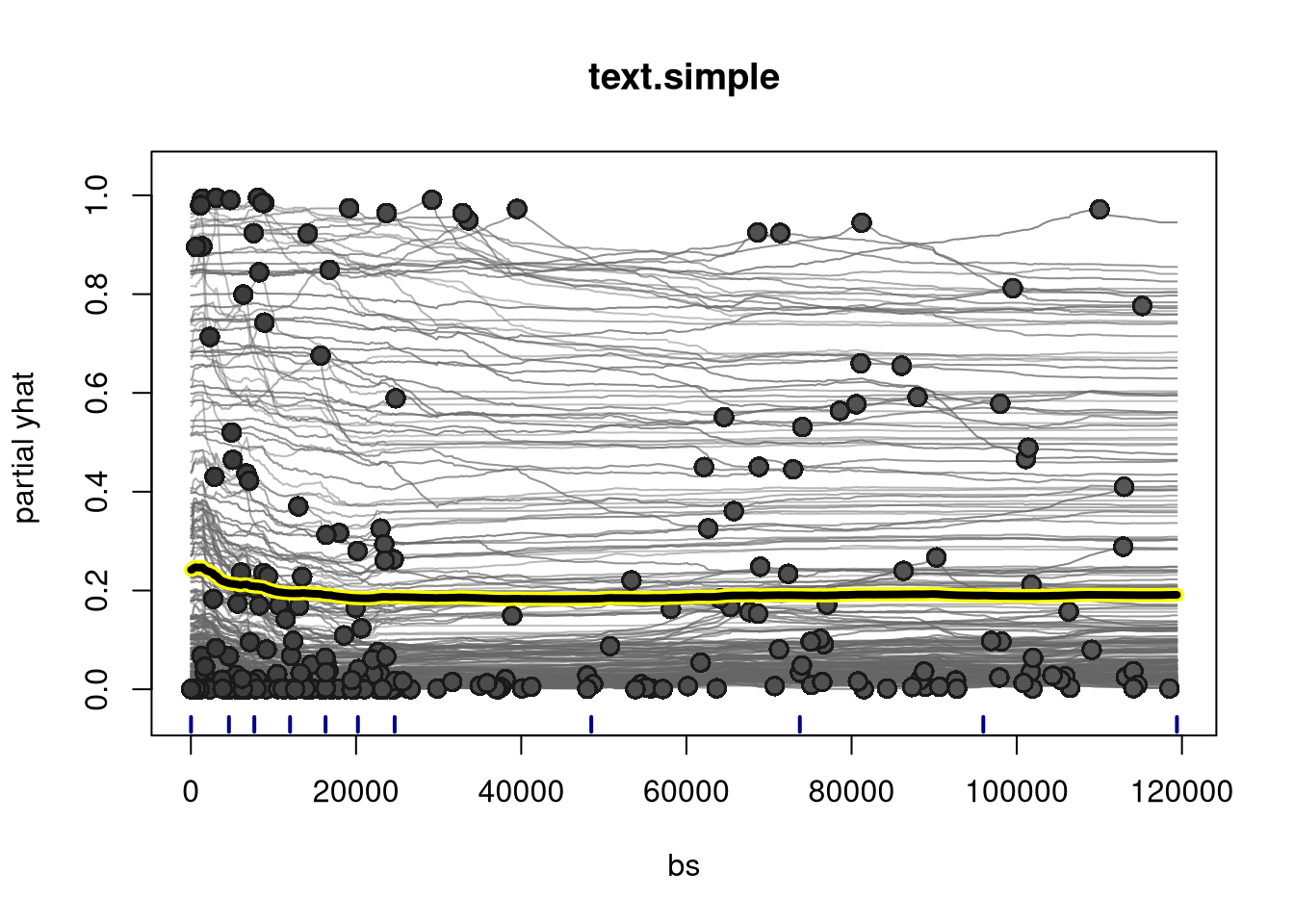}
     \end{subfigure}
     \caption{ICE-plots for the batch size across all three experiment sets}
     \label{fig:ice}
\end{figure*}

\begin{figure*}[t!]
     \begin{subfigure}[b]{0.32\textwidth}
         \centering
         \includegraphics[width=\textwidth]{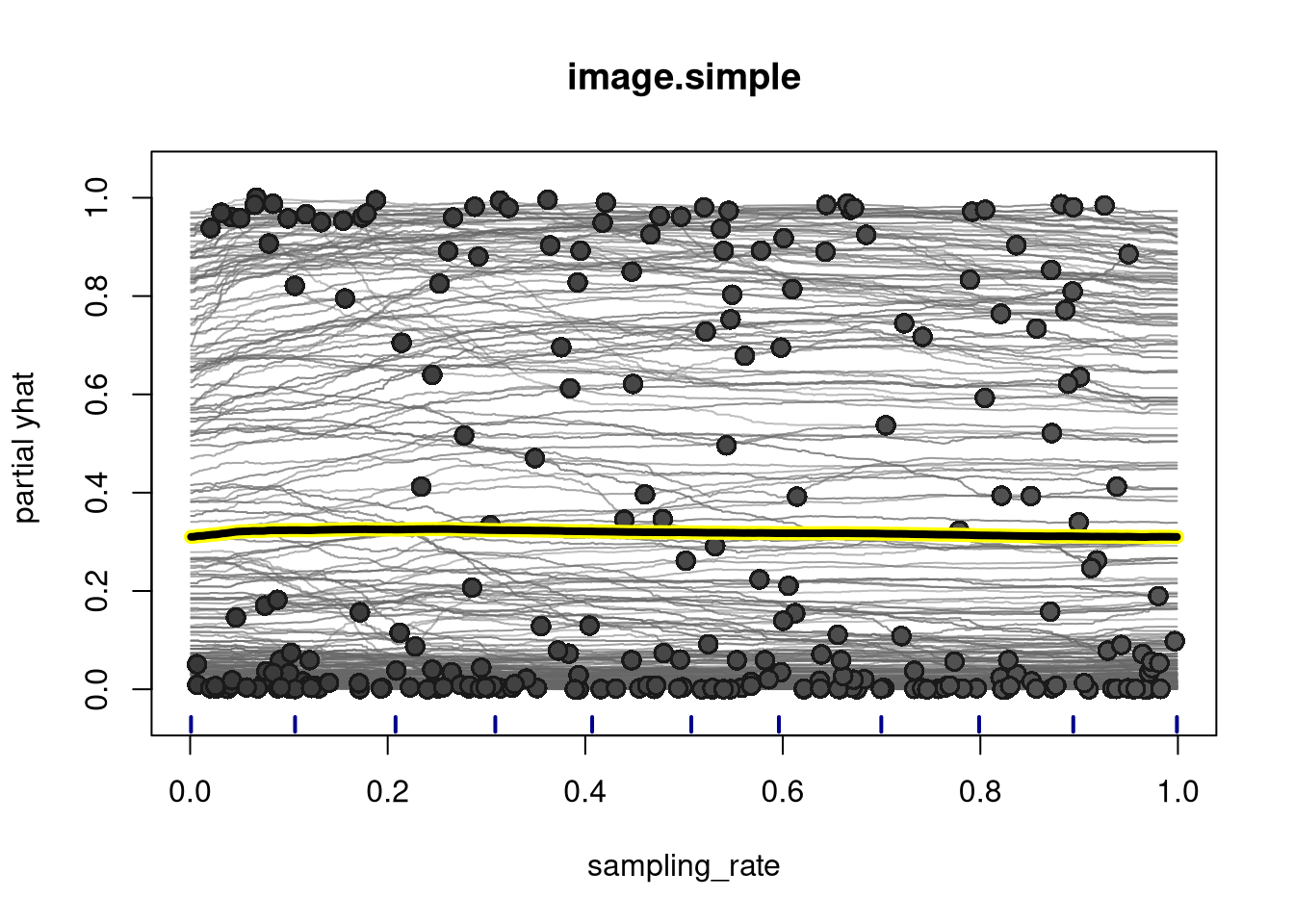}
     \end{subfigure}
     \hfill
     \begin{subfigure}[b]{0.32\textwidth}
         \centering
         \includegraphics[width=\textwidth]{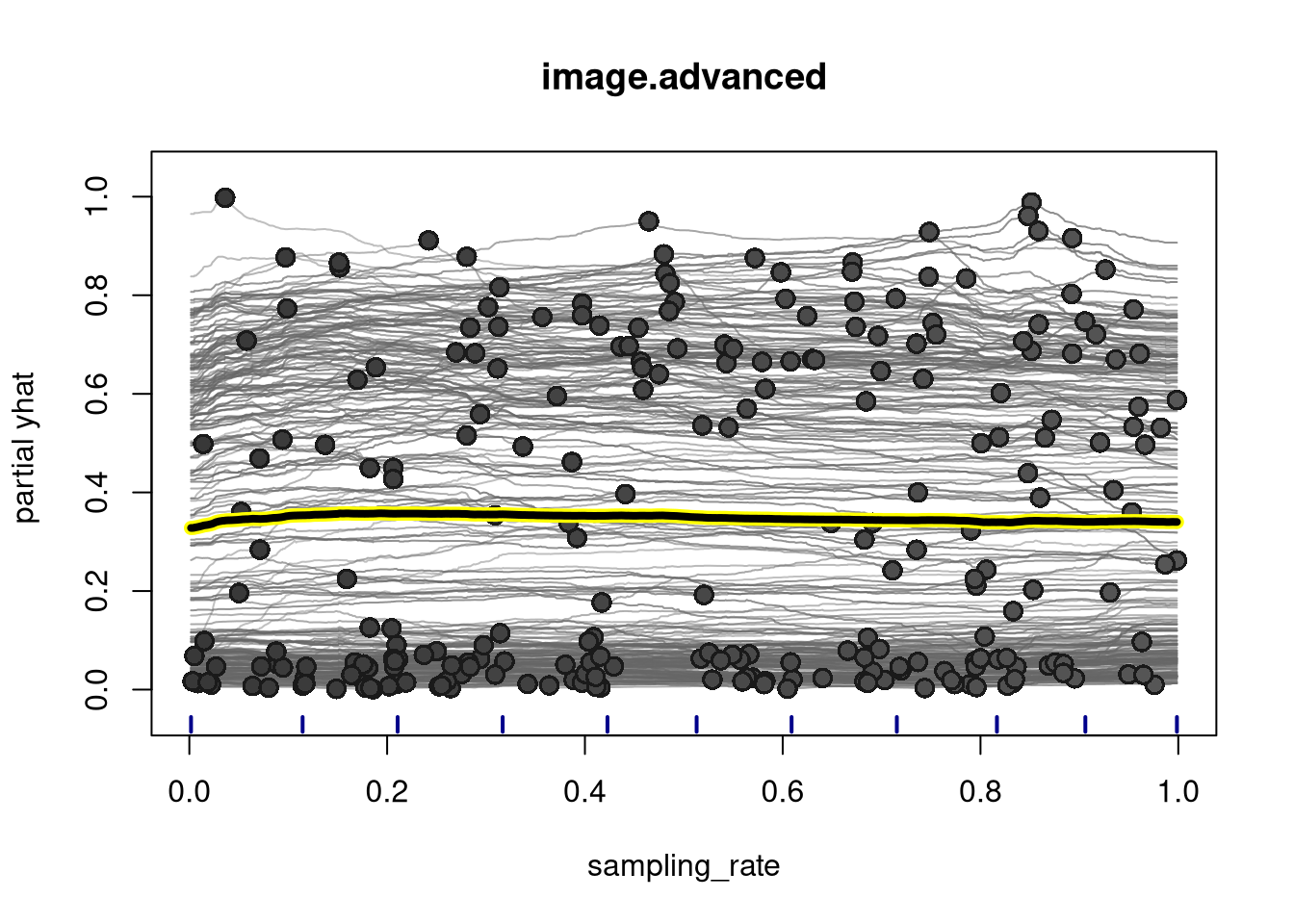}
     \end{subfigure}
     \hfill
     \begin{subfigure}[b]{0.32\textwidth}
         \centering
         \includegraphics[width=\textwidth]{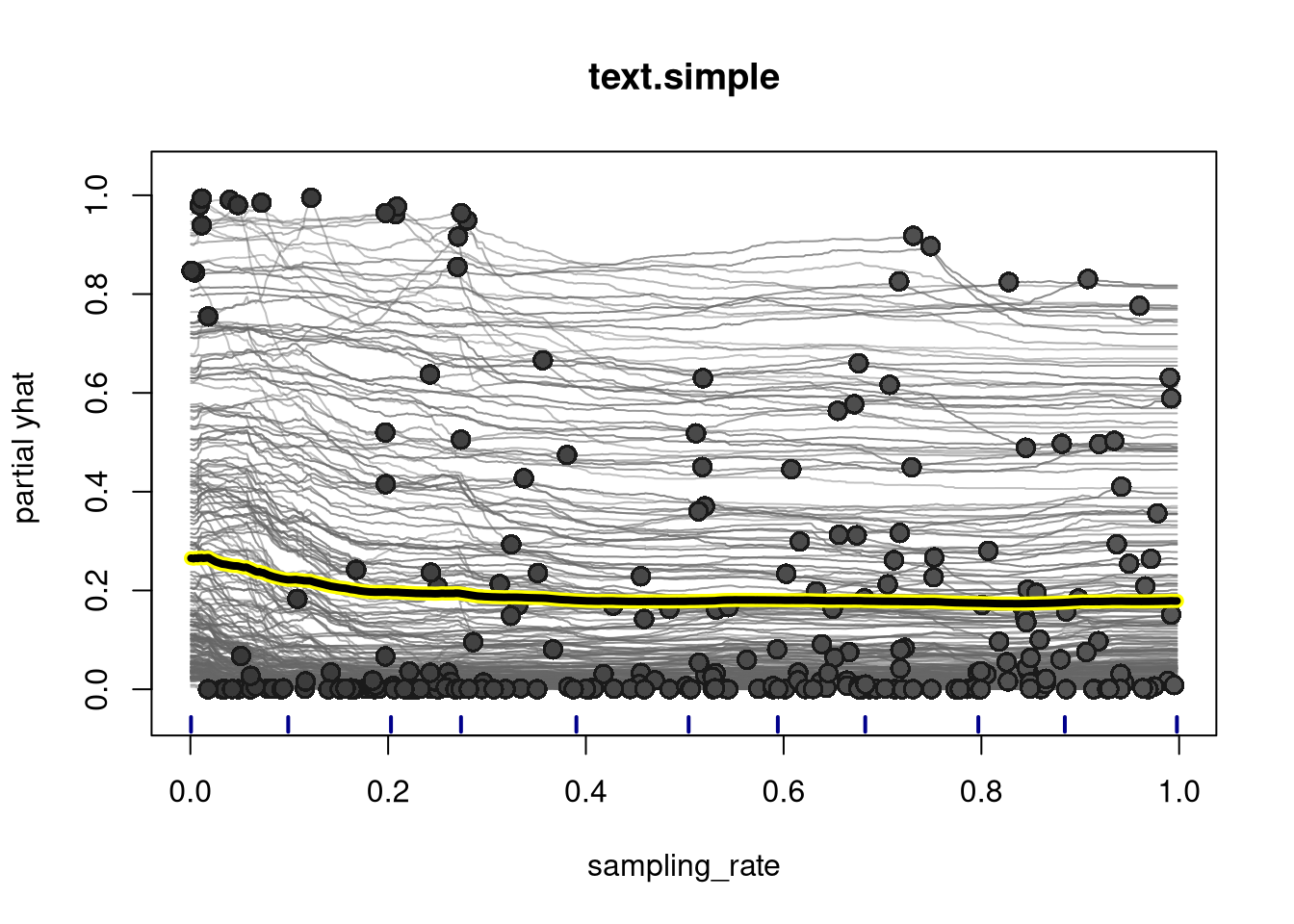}
     \end{subfigure}
     \begin{subfigure}[b]{0.32\textwidth}
         \centering
         \includegraphics[width=\textwidth]{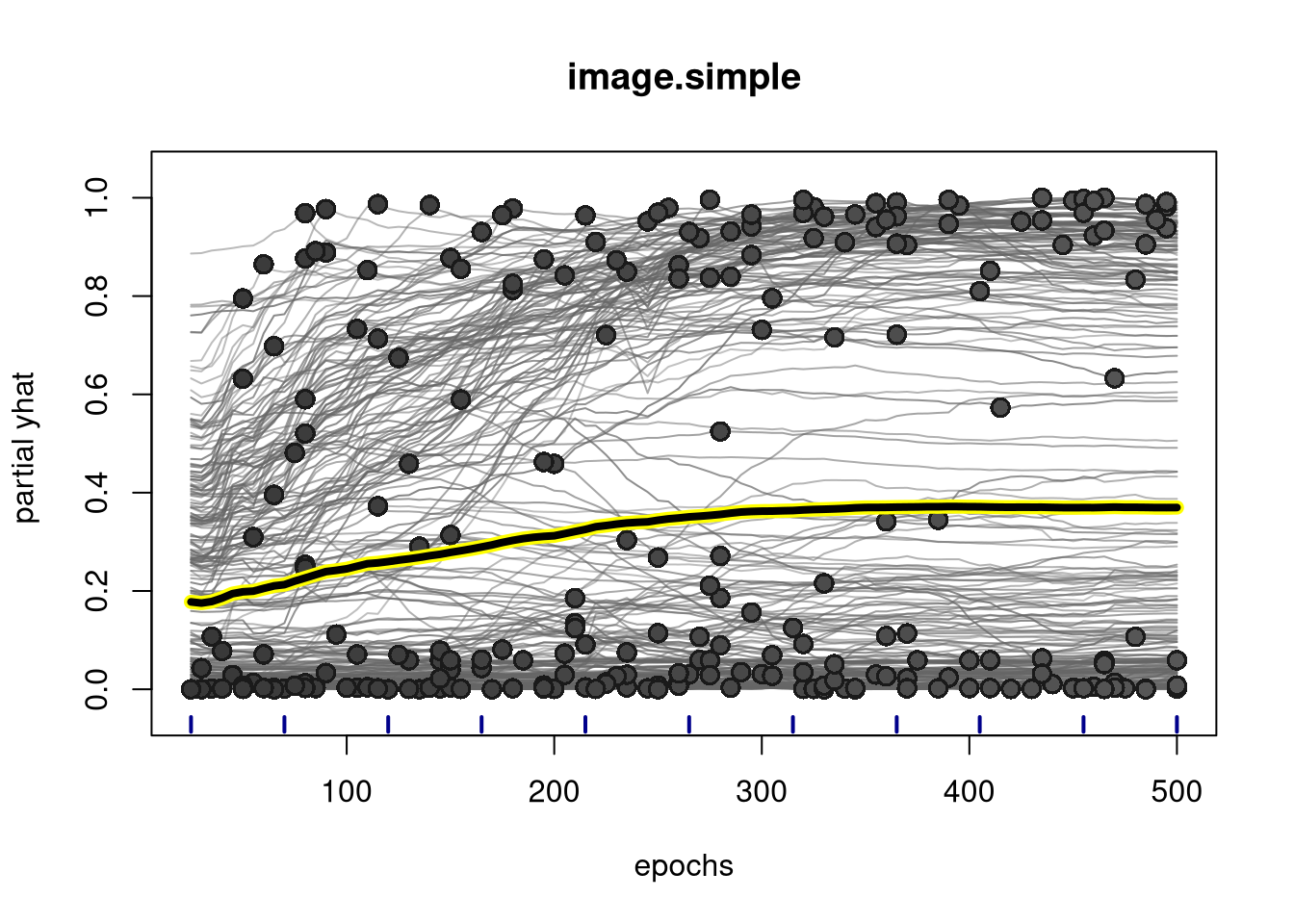}
        \end{subfigure}
     \hfill
     \begin{subfigure}[b]{0.32\textwidth}
         \centering
         \includegraphics[width=\textwidth]{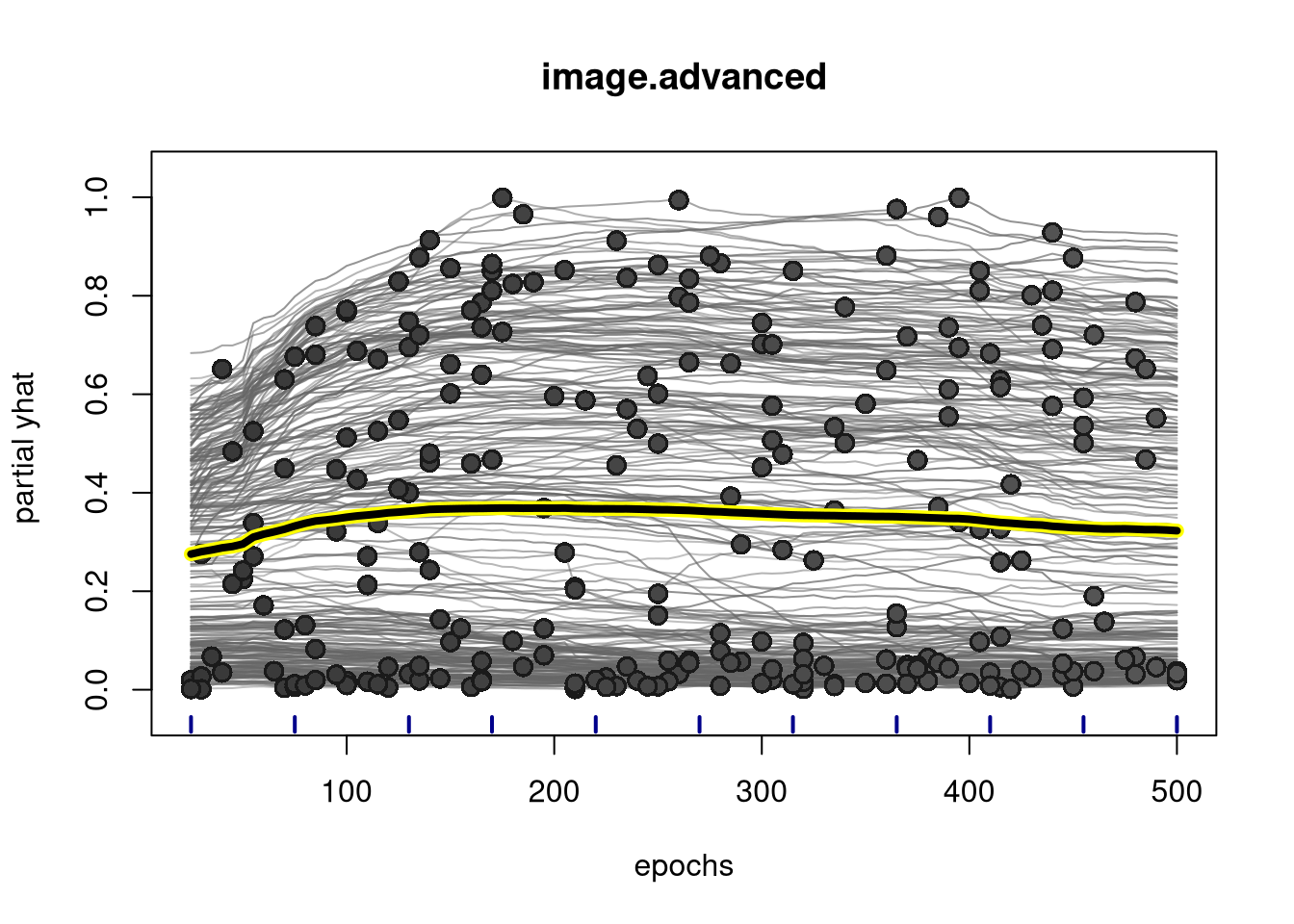}
     \end{subfigure}
     \hfill
     \begin{subfigure}[b]{0.32\textwidth}
         \centering
         \includegraphics[width=\textwidth]{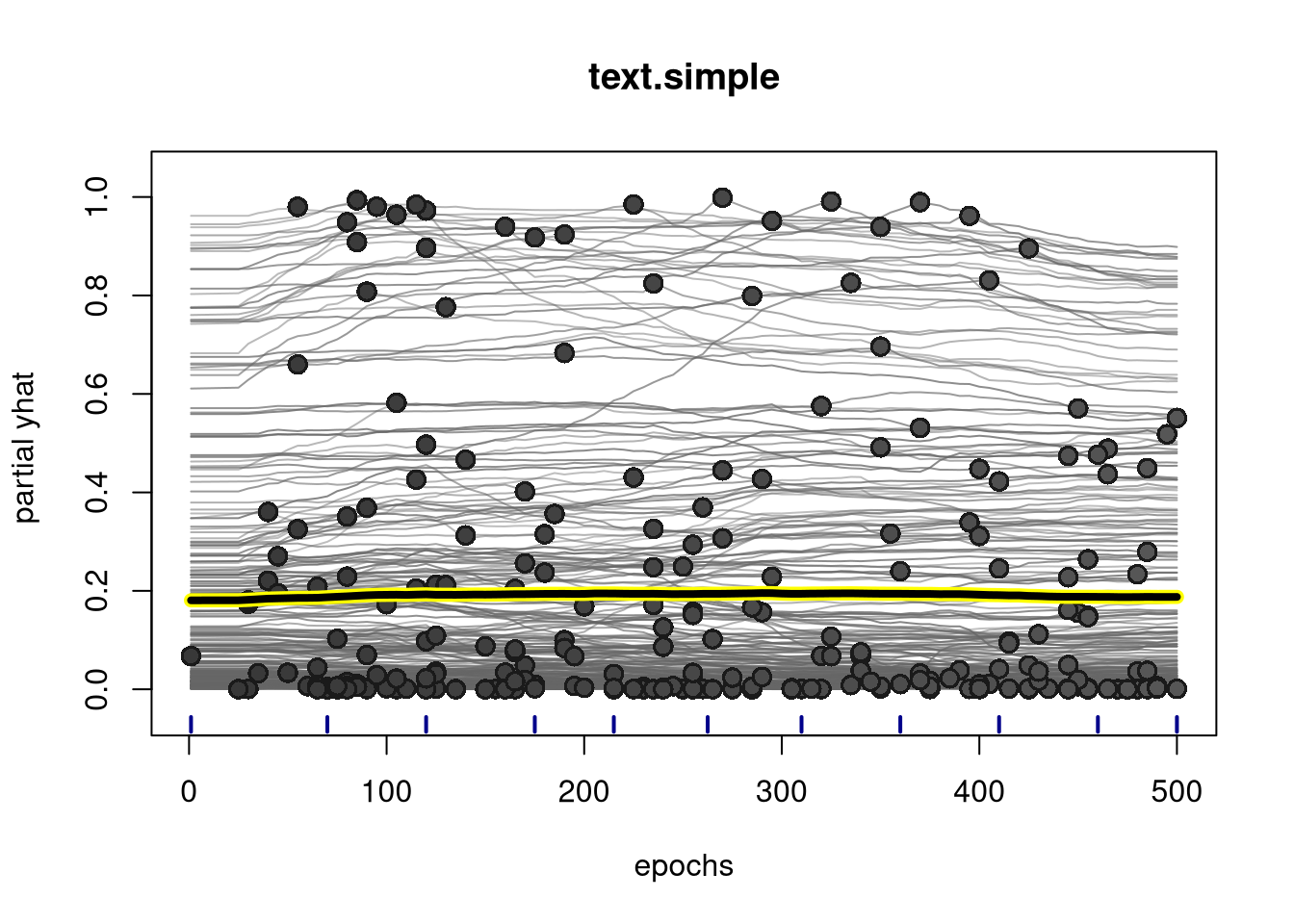}
     \end{subfigure}
     \begin{subfigure}[b]{0.32\textwidth}
         \centering
         \includegraphics[width=\textwidth]{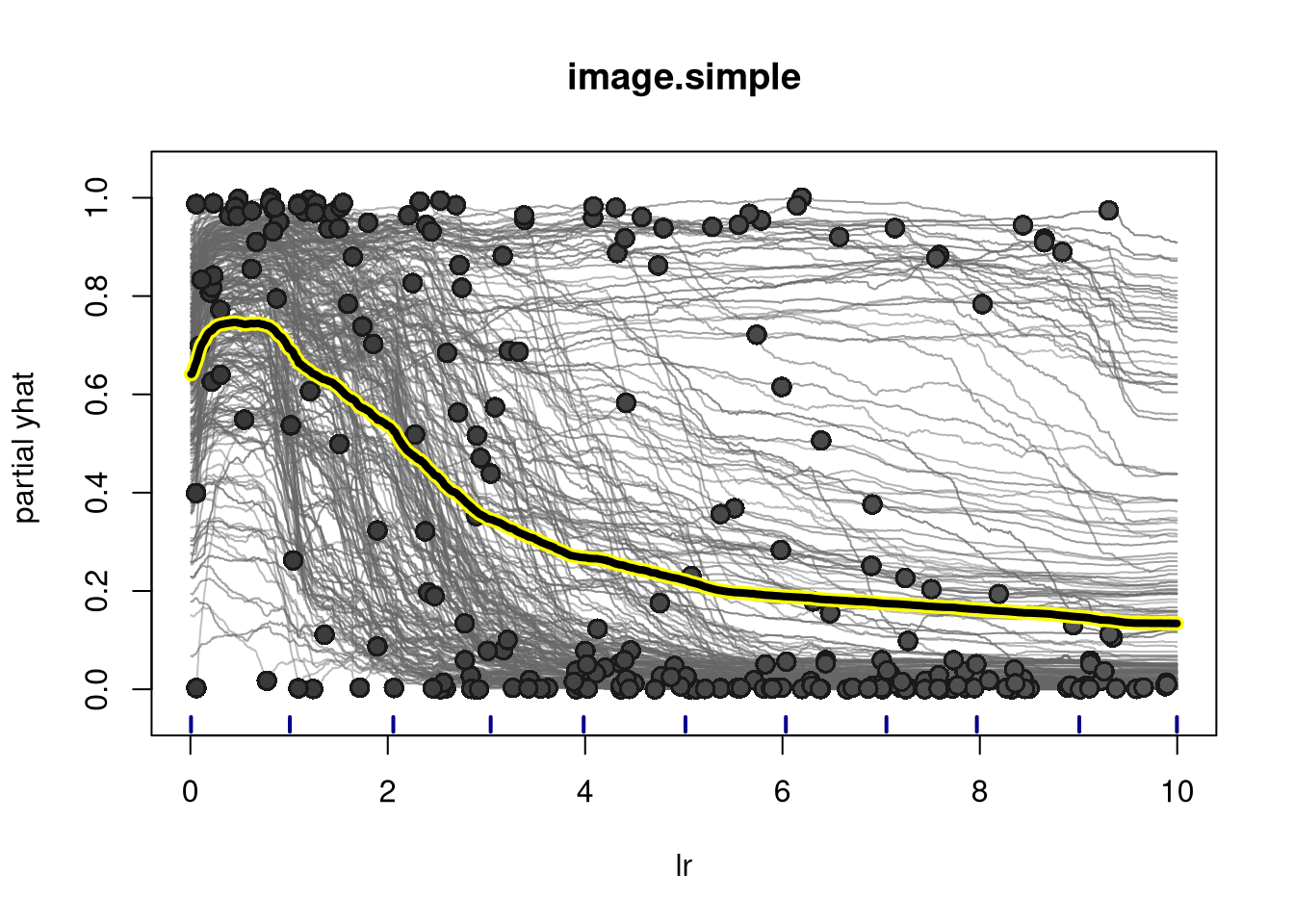}
     \end{subfigure}
     \hfill
     \begin{subfigure}[b]{0.32\textwidth}
         \centering
         \includegraphics[width=\textwidth]{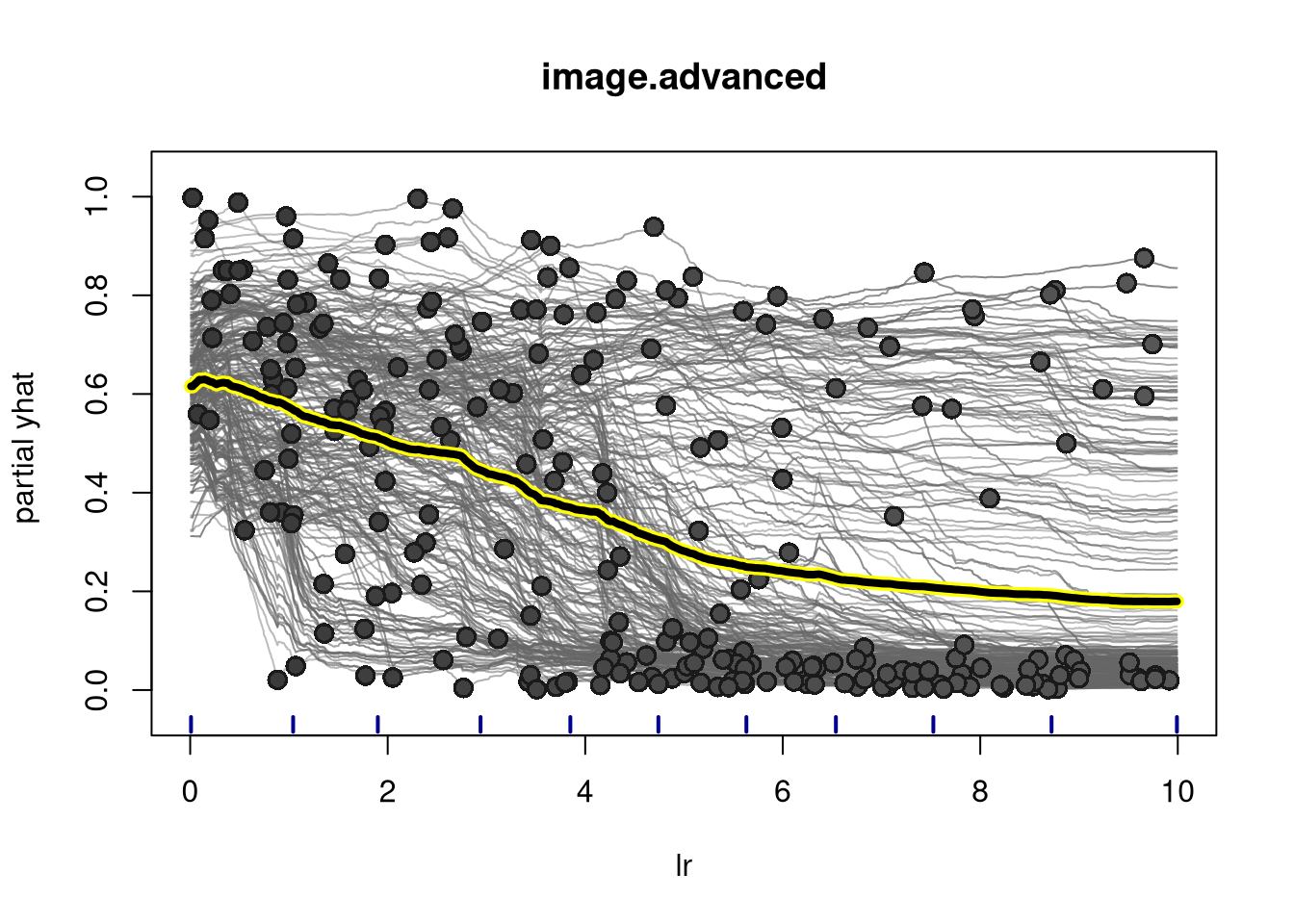}
     \end{subfigure}
     \hfill
     \begin{subfigure}[b]{0.32\textwidth}
         \centering
         \includegraphics[width=\textwidth]{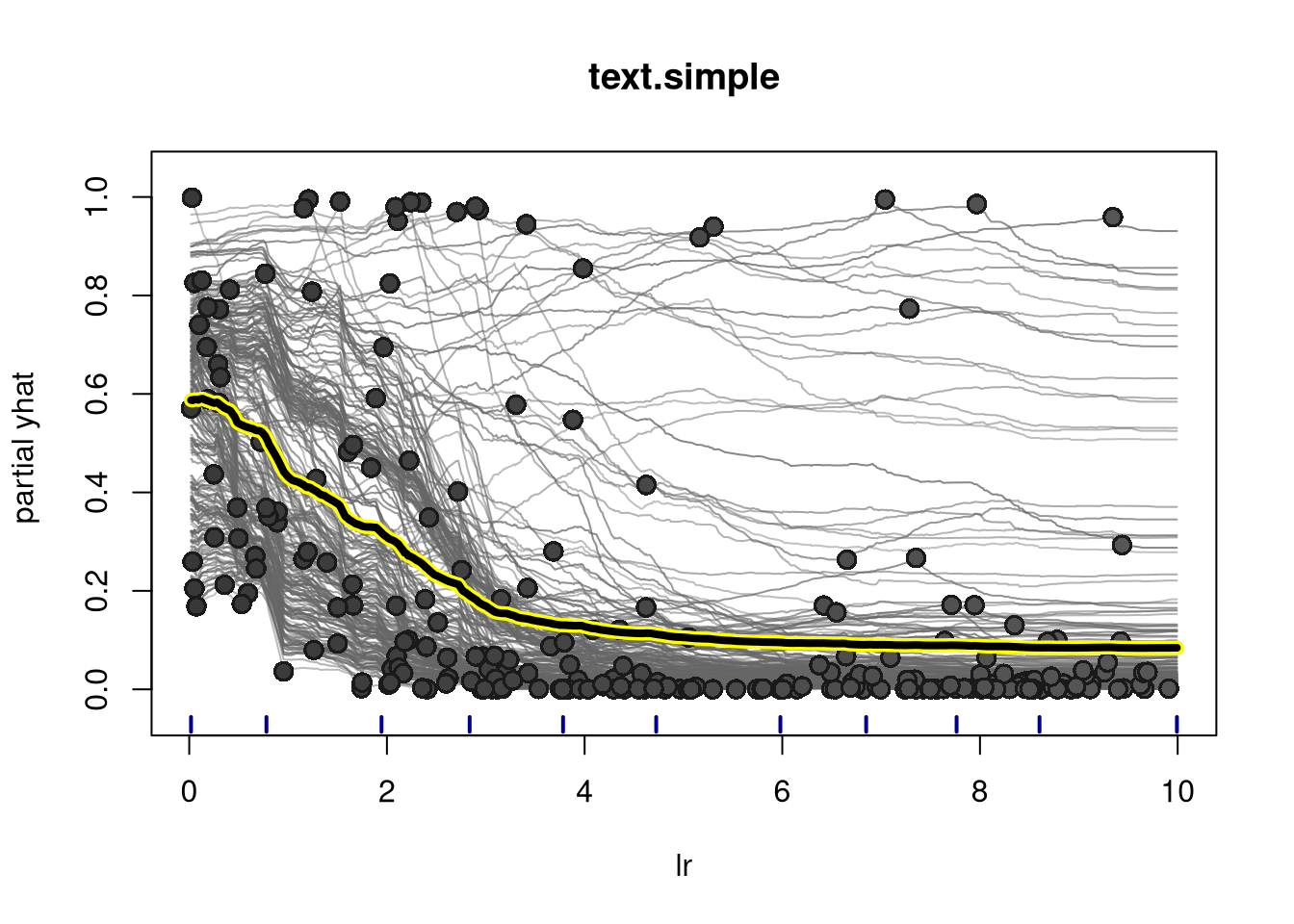}
     \end{subfigure}
     \begin{subfigure}[b]{0.32\textwidth}
         \centering
         \includegraphics[width=\textwidth]{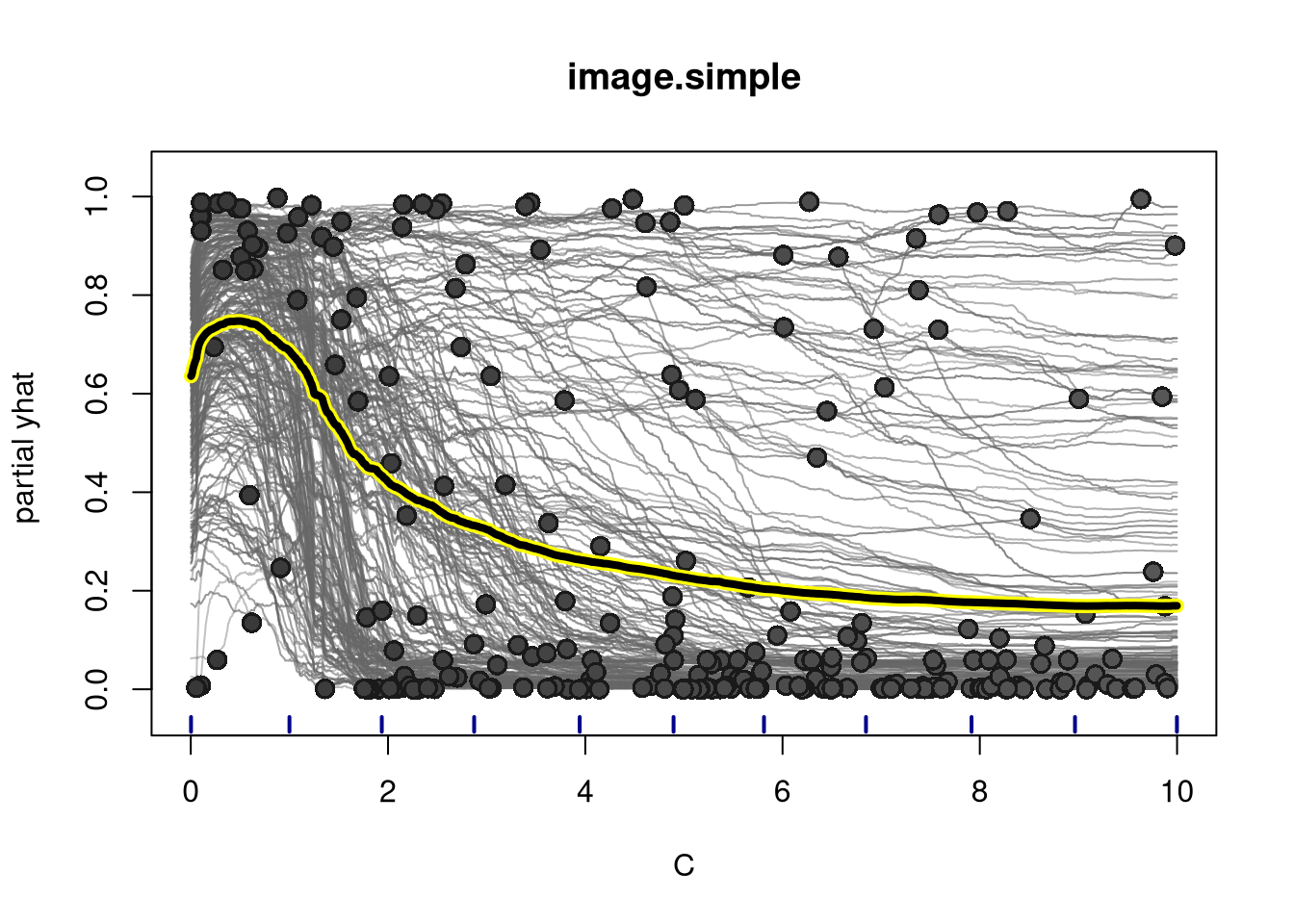}
     \end{subfigure}
     \hfill
     \begin{subfigure}[b]{0.32\textwidth}
         \centering
         \includegraphics[width=\textwidth]{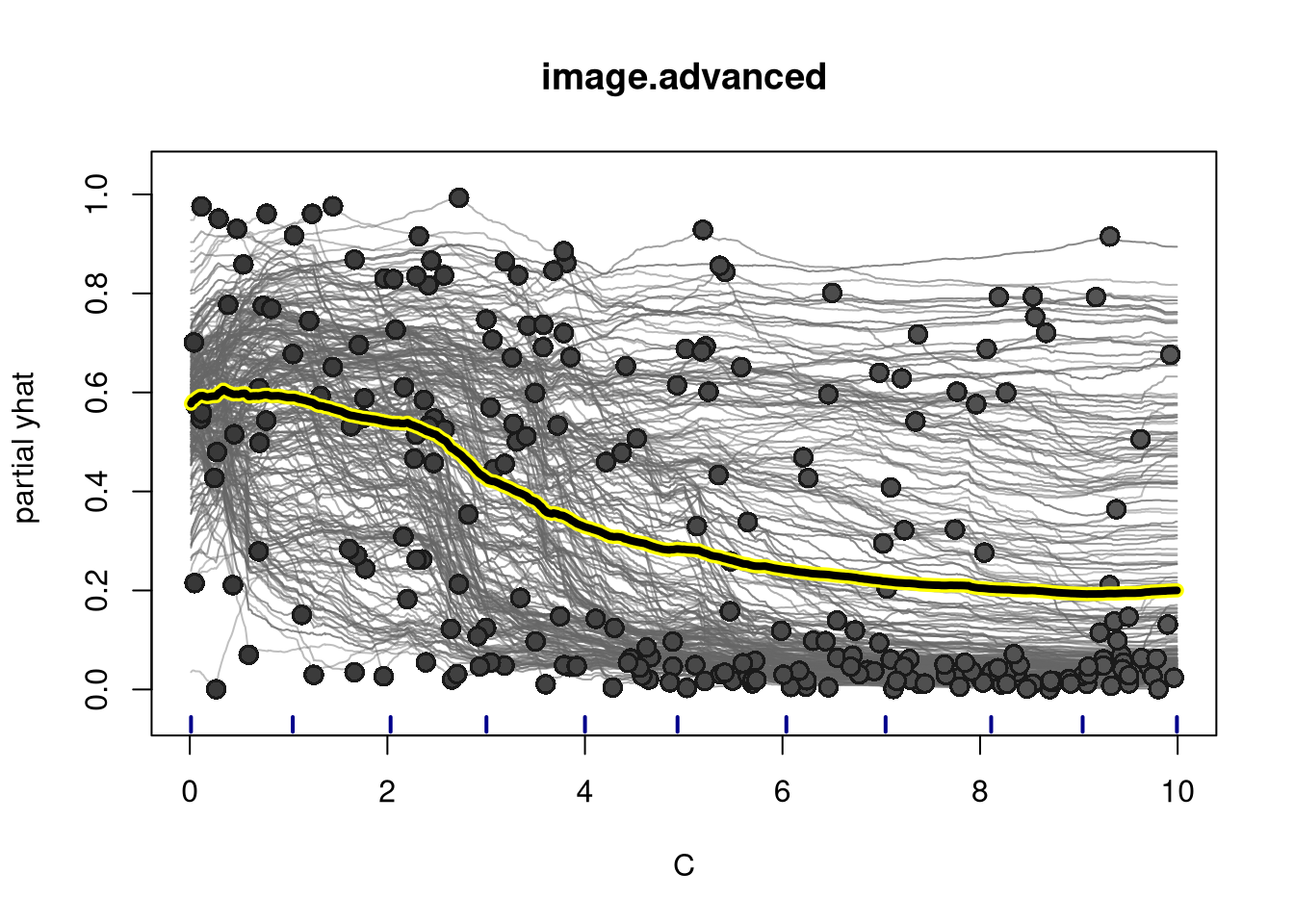}
     \end{subfigure}
     \hfill
     \begin{subfigure}[b]{0.32\textwidth}
         \centering
         \includegraphics[width=\textwidth]{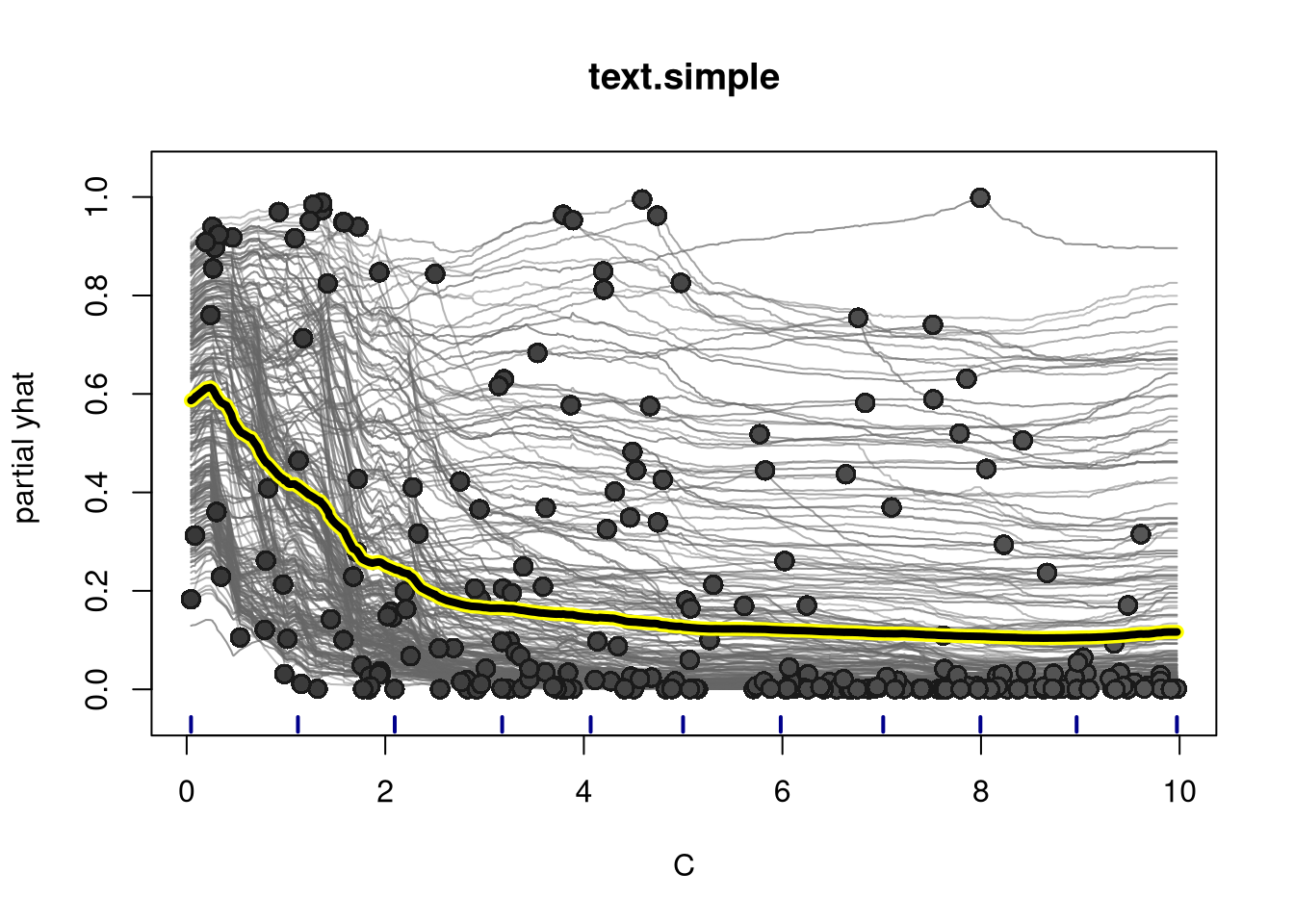}
     \end{subfigure}
    \caption{ICE-plots for sampling rate, number of epochs, learning rate, and clipping threshold across all three experiment sets}
    \label{fig:ice2}
\end{figure*}

\appendix
\label{sec:further_details}
Please find below c-ICE and d-ICE plots on the interaction effects of the clipping threshold and learning rate (Figure \ref{fig:c_d_ice_C_by_lr}), as well as c-ICE and 2d ALE plots (Figure \ref{fig:c-ice_2d-ale_epochs_C}) for the interaction of the number of epochs and clipping threshold. 
Both have been omitted from the main part for brevity. 
For completeness, we also include the individual conditional expectation plots of all hyperparameters across all three experiment sets below, see Figure \ref{fig:ice} and \ref{fig:ice2}. 
Please note that for visual clarity, we only plotted a random fraction of all ICE curves.

\end{document}